\RequirePackage{fix-cm}
\documentclass[%
    twoside, openright, titlepage, numbers=noenddot,%
    cleardoublepage=empty,%
    abstractoff,%
    BCOR=15mm, paper=a4, fontsize=10pt,% A5 soft cover
]{scrreprt}
\usepackage{etex}
\reserveinserts{10}

\usepackage[utf8]{inputenc}

\usepackage[main=english,ngerman]{babel}

\PassOptionsToPackage{%
  eulerchapternumbers,
  floatperchapter, pdfspacing,%
  beramono,%
}{classicthesis}

\newcommand{\myTitle}{%
Analog Alchemy: Neural Computation with In-Memory Inference, Learning and Routing\xspace%
}
\newcommand{\myPlainTitle}{%
Analog Alchemy: Neural Computation with In-Memory Inference, Learning and Routing\xspace%
}

\newcommand{\myName}{Yigit Demirag\xspace}

\newcommand{\myTime}{2024\xspace}

\usepackage{scrtime}

\newcounter{dummy}

\usepackage{csquotes}  % smart quotes
\usepackage[T1]{fontenc}
\usepackage{xspace}
\usepackage{textcomp}
\usepackage{mparhack}
\usepackage{relsize}

\usepackage[
  style=nature,%
  natbib=true,%
  clearlang=true,%
  backend=biber,%
]{biblatex}

\AtEveryBibitem{%
  \clearfield{day}%
  \clearfield{month}%
  \clearfield{endday}%
  \clearfield{endmonth}%
}

\DeclareRedundantLanguages{en,EN,English}{english}

\DeclareFieldFormat{pages}{\mkfirstpage{#1}}

\let\origcite\cite%
\def\cite#1{\unskip~\origcite{#1}}

\usepackage{amsmath}
\usepackage{mathtools}
\usepackage{isomath}

\usepackage{algorithmic}
\usepackage[ruled,vlined]{algorithm2e}

\usepackage{tabularx}
\usepackage{ltablex}
\usepackage{floatrow}
\usepackage{caption}
\usepackage{subcaption}
\usepackage{wrapfig}
\usepackage{blindtext}

\usepackage[dvipsnames]{xcolor}
\usepackage[%
  hyperfootnotes=false,%
  pdfpagelabels,%
]{hyperref}
\usepackage{hyperxmp}
\hypersetup{%
  colorlinks=true, linktocpage=true,%
  breaklinks=true, pdfpagemode=UseNone, pageanchor=true, pdfpagemode=UseOutlines,%
  plainpages=false, bookmarksnumbered, bookmarksopen=true, bookmarksopenlevel=1,%
  hypertexnames=true, pdfhighlight=/O,%nesting=true,%frenchlinks,%
  pdftitle={\myPlainTitle},%
  pdfauthor={\myName},%
  pdfcopyright={Copyright (C) \myTime, \myName},%
  pdfsubject={},%
  pdfkeywords={},%
  pdflang={en},%
}

\usepackage{graphicx}
\usepackage{rotating}
\usepackage{tikz}
\usepackage{pgfplots}
\usepackage{multirow}
\pgfplotsset{compat=newest}
\usetikzlibrary{shapes.geometric}

\usepackage{cleveref}

\PassOptionsToPackage{printonlyused}{acronym}
\usepackage{acronym}

\usepackage{enumitem}

\usepackage{scrhack}
\usepackage{classicthesis}
\usepackage[a4paper, inner=3.5cm, outer=2.5cm, top=3cm, bottom=3cm]{geometry}

\areaset[current]{\textwidth}{1.618034\textwidth}

\clearscrplain
\ofoot[\pagemark]{}
\definecolor{chapter-color}{cmyk}{1, 0.50, 0, 0.25}
\definecolor{link-color}{cmyk}{1, 0.50, 0, 0.25}
\definecolor{cite-color}{cmyk}{0, 0.7, 0.9, 0.2}

\usepackage{bookmark}
\hypersetup{
  urlcolor=Black, linkcolor=Black, citecolor=Black, %pagecolor=Black,%
}

\let\chapterNumber\undefined%
\newfont{\chapterNumber}{eurb10 scaled 5500}
\usepackage{siunitx}
\DeclareSIUnit\au{a.u.}
\titleformat{\chapter}[display]%
  {\relax}{\vspace*{-3\baselineskip}\makebox[\linewidth][r]{\color{halfgray}\chapterNumber\thechapter}}{10pt}%
  {\raggedright\spacedallcaps}[\normalsize\vspace*{.8\baselineskip}\titlerule]%

\setkomafont{dictumtext}{\itshape\small}%
\setkomafont{dictumauthor}{\normalfont}
\usepackage[final]{pdfpages}

\begin{document}
\frenchspacing
\raggedbottom%
\selectlanguage{english}
\pagenumbering{roman}
\pagestyle{scrplain}

%
% Cover
%
% Uncomment and adapt these lines if you want to include a cover PDF.
%
%\includepdf[pages={1,{}}]{cover/crop/cover_front.pdf}
\includepdf[pages={1}]{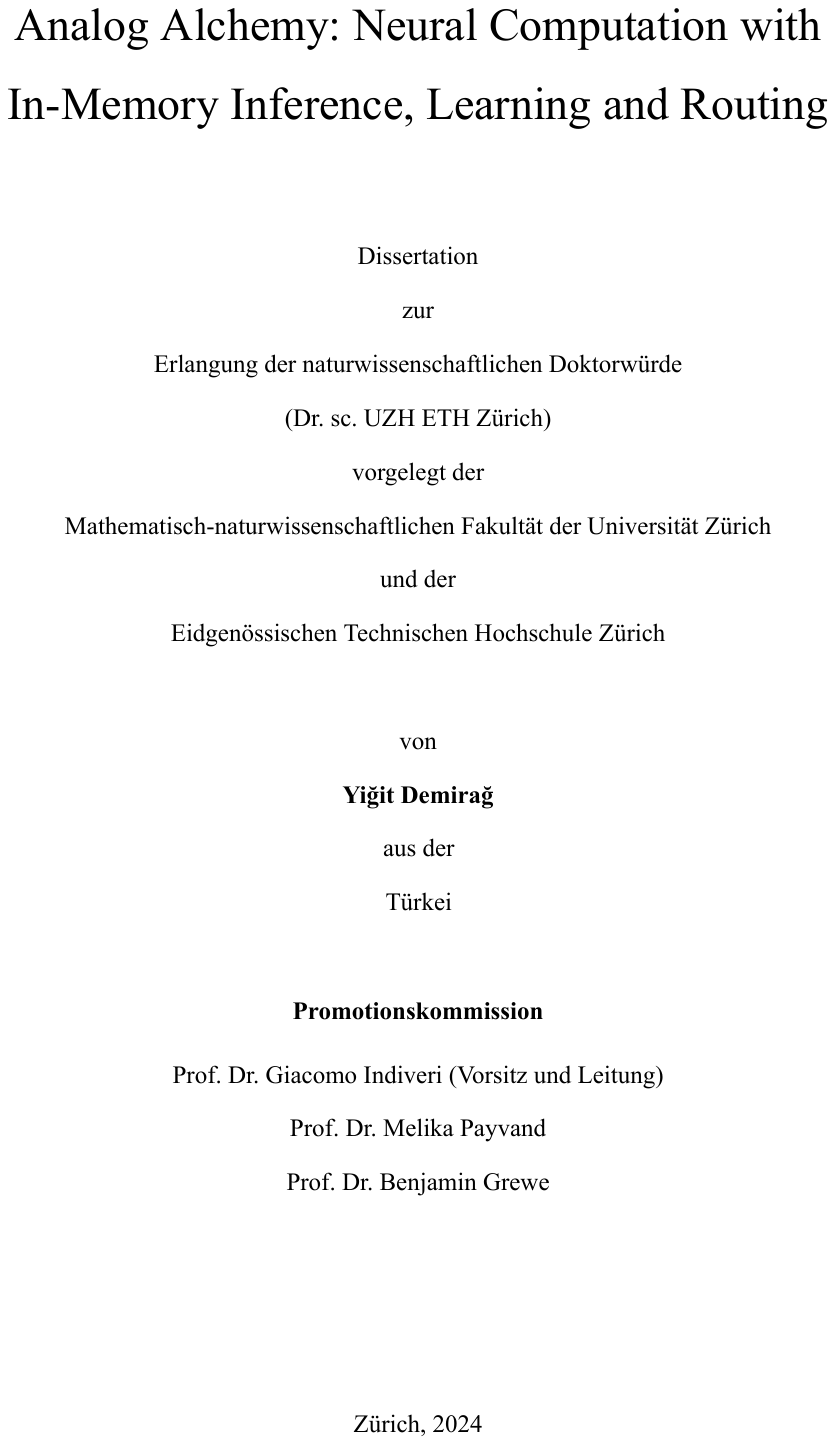}
\cleardoublepage\setcounter{page}{1}

%
% Frontmatter
%

\cleardoublepage%*******************************************************
% Little Dirty Titlepage
%*******************************************************
\thispagestyle{empty}
%*******************************************************
\begin{center}
    \spacedlowsmallcaps{\myName} \\ \medskip                        

    \begingroup
        \color{chapter-color}\spacedallcaps{\myTitle}
    \endgroup
\end{center}        

%\cleardoublepage\include{frontbackmatter/titlepage}
%\include{frontbackmatter/titleback}
\cleardoublepage%*******************************************************
% Dedication
%*******************************************************
\thispagestyle{empty}
%\phantomsection
\refstepcounter{dummy}
%\pdfbookmark[1]{Dedication}{Dedication}

\vspace*{3cm}

\begin{center}
    To the engineers and scientists\\
    who will one day build superintelligence;\\
    from whatever materials and circuits,\\
    in whatever form.
\end{center}

\medskip

\cleardoublepage%*******************************************************
% Abstract
%*******************************************************
%\renewcommand{\abstractname}{Abstract}
\pdfbookmark[1]{Abstract}{Abstract}
\begingroup
\let\clearpage\relax
\let\cleardoublepage\relax
\let\cleardoublepage\relax

\chapter*{Abstract}

% problem
As neural computation is revolutionizing the field of \ac{AI}, rethinking the ideal neural hardware is becoming the next frontier.
Fast and reliable von Neumann architecture has been the hosting platform for neural computation.
Although capable, its separation of memory and computation creates the bottleneck for the energy efficiency of neural computation, contrasting the biological brain.
The question remains: how can we efficiently combine memory and computation, while exploiting the physics of the substrate, to build intelligent systems?
% approach
In this thesis, I explore an alternative way with memristive devices for neural computation, where the unique physical dynamics of the devices are used for inference, learning and routing.
Guided by the principles of gradient-based learning, we selected functions that need to be materialized, and analyzed connectomics principles for efficient wiring.
Despite non-idealities and noise inherent in analog physics, I will provide hardware evidence of adaptability of local learning to memristive substrates, new material stacks and circuit blocks that aid in solving the credit assignment problem and efficient routing between analog crossbars for scalable architectures.
% methods
First, I address limited bit precision problem in binary \ac{RRAM} devices for stable training.
By introducing a new device programning technique that precisely controls the filament growth process, we enhance the effective bit precision of these devices.
Later, we prove the versatility of this technique by applying it to novel perovskite memristors.
Second, I focus on the hard problem of online credit assignment in recurrent \acp{SNN} in the presence of memristor non-idealities.
I present a simulation framework based on a comprehensive statistical model of \ac{PCM} crossbar array, capturing all major device non-idealities. 
Building upon the recently developed e-prop local learning rule, we demonstrate that gradient accumulation is crucial for reliably implementing the learning rule with memristive devices.
Moreover, I introduce PCM-trace, a scalable implementation of synaptic eligibility traces, a functional block demanded by many learning rules, using volatile characteristics by specifically fabricated \ac{PCM} devices.
Third, I present our discovery of a novel memristor material capable of switching between volatile and non-volatile modes. 
This reconfigurable memristor, based on halide perovskite nanocrystals, offers a significant advancement in emerging memory technologies, enabling the implementation of both static and dynamic neural variables with the same material and fabrication technique, while holding the world record in endurance.
Finally, I introduce Mosaic, a memristive systolic architecture for in memory computing and routing. 
Mosaic, trained with our novel layout-aware training methods, efficiently implements small-world graph connectivity and demonstrates superior energy efficiency in spike routing compared to other hardware platforms.
\endgroup

\vfill
\cleardoublepage%*******************************************************
% Acknowledgments
%*******************************************************
\pdfbookmark[1]{Acknowledgements}{acknowledgements}

\bigskip

\begingroup
\let\clearpage\relax
\let\cleardoublepage\relax
\let\cleardoublepage\relax
\chapter*{Acknowledgements}

\def\thanks#1{%
\begingroup
\leftskip1em
\noindent #1
\par
\endgroup
}

This thesis wouldn't have been possible without many people: scientists, friends, and family. I'm honored to have shared this journey with such curious and driven individuals. Among the many who contributed, there are a few exceptional individuals who were absolutely core to making this happen:\\

First I'd like to thank my supervisor, Giacomo Indiveri, who is a rare scientist truly channeling his work towards a dream.
Over these 5 years, he gave me complete freedom to explore what I believe are the most exciting problems, while providing me with high-bandwidth feedback on demand, with more than 500 emails and many thousands of DMs.
He taught me the importance of pushing unusual ideas to the limit.
Whenever I came up with an ambitious project goal, he always reminded me to deeply consider the efficiency on the silicon, first.
I'm grateful for having been his student.\\

I've been very fortunate to coincide with Melika Payvand, my co-supervisor, in this particular academic space and time.
She is the most curious mind craving to understand the emergence of intelligence from the physics of computation, and her passion is infectious.
Together, we traversed the probability trees for nearly every projects in my PhD, and executed against the entropy.
Her close friendship is the cherry on the cake; I enjoyed and valued every second of it.\\

Then there are people I am very lucky to collaborate with and learn from.
Rohit A. John, an extraordinary person who taught me the importance of grinding with massive focus while solving hard problems.
And Elisa Vianello, who always provided her seamless support and insights that have made hard projects a joyful exploration.
And Emre Neftci, whose disruptive scientific ideas deeply resonated with me, and with whom I always enjoyed discussing ideas.\\

I have to thank Alpha, Anqchi, Arianna, Chiara, Dmitrii, Farah, Filippo, Jimmy, Karthik, Manu, Maryada, Nicoletta, Tristan, and many others, who I hope will forgive me for not being mentioned individually or for resorting to alphabetical order when I did. Thank you for inspiring conversations in INI hallways, night walks in Zurich, and giving me the privilege of calling you, my friends.\\

During my PhD, I completed two internships at Google Zurich and one research visit at MILA.
All of these were fantastic learning experiences, where I got a chance to reshape my research scope.
From these experiences, I would especially like to thank Jyrki Alakuijala, Johannes von Oswald, Eyvind Niklasson, Ettore Randazzo, Alexander Mordvintsev, Esteban Real, Arna Ghosh, Jonathan Cornford, Joao Sacramento, Blake Richards, Guillaume Lajoie, and Blaise Aguera y Arcas.\\

I would also like to extend my thanks to my professors from my Master's degree, particularly Ekmel Ozbay, Bayram Butun and Yusuf Leblebici, for their invaluable support and inspiration.\\

And to Gizay, for sharing most of the journey and everything we created together.\\

But most of all, I want to express my deepest gratitude to my mom and dad, who inspired me to be curious, to take the world as a playground, and provided me with a loving home.
And to my little brother, Efe, who is the best teammate in every game we play and in life's adventures.
\endgroup

\pagestyle{scrheadings}
\cleardoublepage%*******************************************************
% Table of Contents
%*******************************************************
%\phantomsection
\refstepcounter{dummy}
\pdfbookmark[1]{\contentsname}{tableofcontents}
\setcounter{tocdepth}{2} % <-- 2 includes up to subsections in the ToC
\setcounter{secnumdepth}{3} % <-- 3 numbers up to subsubsections
\manualmark%
\markboth{\spacedlowsmallcaps{\contentsname}}{\spacedlowsmallcaps{\contentsname}}
\tableofcontents
\automark[section]{chapter}
\renewcommand{\chaptermark}[1]{\markboth{\spacedlowsmallcaps{#1}}{\spacedlowsmallcaps{#1}}}
\renewcommand{\sectionmark}[1]{\markright{\thesection\enspace\spacedlowsmallcaps{#1}}}
%*******************************************************
% List of Figures and of the Tables
%*******************************************************
\clearpage

\begingroup
    \let\clearpage\relax
    \let\cleardoublepage\relax
    \let\cleardoublepage\relax
    %*******************************************************
    % List of Figures
    %*******************************************************
    %\phantomsection
    % \refstepcounter{dummy}
    %\addcontentsline{toc}{chapter}{\listfigurename}
    % \pdfbookmark[1]{\listfigurename}{lof}
    % \listoffigures

    % \vspace{8ex}

    %*******************************************************
    % List of Tables
    %*******************************************************
    %\phantomsection
    % \refstepcounter{dummy}
    %\addcontentsline{toc}{chapter}{\listtablename}
    % \pdfbookmark[1]{\listtablename}{lot}
    % \listoftables

    % \vspace{8ex}
    % \newpage

    %*******************************************************
    % List of Listings
    %*******************************************************
      %\phantomsection
    %\refstepcounter{dummy}
    %\addcontentsline{toc}{chapter}{\lstlistlistingname}
    %\pdfbookmark[1]{\lstlistlistingname}{lol}
    %\lstlistoflistings%

    %\vspace{8ex}

    % Notation
    %\refstepcounter{dummy}
    %\pdfbookmark[1]{Notation}{notation}
    %\markboth{\spacedlowsmallcaps{Notation}}{\spacedlowsmallcaps{Notation}}
    %\chapter*{Notation}

    %\section*{Frequently used symbols}%
    %\vskip -2em
    %\begin{tabularx}{\textwidth}{lX}
      %\toprule%
      %\tableheadline{Symbol} & \tableheadline{Meaning} \\
      %\midrule%
    %  $E$ & energy \\
    %  $m$ & rest mass \\
    %  $p$ & impulse \\
      %\bottomrule
    %\end{tabularx}

    %\section*{Physical constants}
    %\sisetup{separate-uncertainty=false}
    %\vskip -2em
    %\begin{tabularx}{\textwidth}{lX}
%$c$ & speed of light in vacuum, $c=\u{299792458}{\metre\per\second}$ \\
    %\end{tabularx}
    %\begin{flushright}
    %(CODATA 2014~\cite{codata})
    %\end{flushright}
    %\sisetup{separate-uncertainty=true}
\endgroup

\acrodef{ADC}[ADC]{Analog-to-Digital Converter}
\acrodef{CCO}[CCO]{Current-Controlled Oscillator}
\acrodef{CC}[CC]{Compliance Current}
\acrodef{CMOS}[CMOS]{Complementary Metal-Oxide-Semiconductor}
\acrodef{HCS}[HCS]{High Conductance State}
\acrodef{HRS}[HRS]{High Resistive State}
\acrodef{SNN}[SNN]{Spiking Neural Network}
\acrodef{TTFS}[TTFS]{Time-to-First-Spike}
\acrodef{AER}[AER]{Address-Event Representation}
\acrodef{GST}[GST]{$\text{Ge}_2\text{Sb}_2\text{Te}_5$}
\acrodef{LIF}[LIF]{Leaky Integrate-and-Fire}
\acrodef{PCM}[PCM]{Phase Change Material}
\acrodef{RL}[RL]{Reinforcement Learning}
\acrodef{LB}[LB]{Learning Block}
\acrodef{OSTL}[OSTL]{Online Spatio-Temporal Learning}
\acrodef{RTRL}[RTRL]{Real Time Recurrent Learning}
\acrodef{RNN}[RNN]{Recurrent Neural Network}
\acrodef{ANN}[ANN]{Artificial Neural Network}
\acrodef{MVM}[MVM]{Matrix-Vector Multiplication}
\acrodef{SRAM}[SRAM]{Static Random Access Memory}
\acrodef{BPTT}[BPTT]{Back-Propagation Through Time}
\acrodef{STDP}[STDP]{Spike-Timing Dependent Plasticity}
\acrodef{LSTM}[LSTM]{Long Short-Term Memory}
\acrodef{OSTL}[OSTL]{Online Spatio-Temporal Learning}
\acrodef{LTD}[LTD]{Long-Term Depression}
\acrodef{LTP}[LTP]{Long-Term Potentiation}
\acrodef{FDSOI}[FDSOI]{Fully Depleted Silicon-On-Insulator}
\acrodef{ET}[ET]{Eligibility Trace}
\acrodef{SDSP}[SDSP]{Spike-Driven Synaptic Plasticity}
\acrodef{LRS}[LRS]{Low Resistive State}
\acrodef{LCS}[LCS]{Low Conductive State}
\acrodef{WTA}[WTA]{Winner-Take-All}
\acrodef{RRAM}[RRAM]{Resistive Random Access Memory}
\acrodef{FeRAM}[FeRAM]{Ferroelectric Random Access Memory}
\acrodef{RSNN}[RSNN]{Recurrent Spiking Neural Network}
\acrodef{ECG}[ECG]{Electrocardiogram}
\acrodef{SHD}[SHD]{Spiking Heidelberg Digits}
\acrodef{CAM}[CAM]{Content Addressable Memory}
\acrodef{ES}[ES]{Evolutionary Strategies}
\acrodef{DPI}[DPI]{Differential-Pair Integrator}
\acrodef{TD}[TD]{Temporal Difference}
\acrodef{AI}[AI]{Artificial Intelligence}
\acrodef{FP32}[FP32]{Single-precision floating-point format}
\acrodef{APMOM}[APMOM]{Alternate Polarity Metal On Metal}
\acrodef{HfO2}[HfO2]{Hafnium oxide}
\acrodef{STP}[STP]{Short-Term Plasticity}
\acrodef{FLOPs}[FLOPs]{Floating point operations per second}
\acrodef{SSM}[SSM]{State Space Model}
\acrodef{TPU}[TPU]{Tensor Processing Unit}
\acrodef{LLM}[LLM]{Large Language Model}
\acrodef{DRAM}[DRAM]{Dynamic Random Access Memory}
\acrodef{SRAM}[SRAM]{Static Random Access Memory}
\acrodef{HBM}[HBM]{High Bandwidth Memory}

%
% Mainmatter
%
\cleardoublepage\pagenumbering{arabic}%
\def\dir{chapters/0_introduction}
\chapter{Appendix}
\label{ch:appendix}

We present here some additional results to complement the main results discussed in the previous sections.

\section*{Appendix 1: Online Training of Spiking Recurrent Neural Networks with Phase-Change Memory Synapses}

\paragraph{Supplementary Note 1}
\label{section:xbar}

We implemented the \ac{PCM} crossbar array simulation framework in PyTorch~\cite{Paszke_etal19a}, which can be used for both the inference and the training of \acp{ANN} or \acp{SNN}.
Built on top of the statistical model introduced by Nandakumar et al.~\cite{Nandakumar_etal18}, our crossbar model supports asynchronous SET, RESET and READ operations over entire crossbar structures and simultaneously keep tracks of the temporal evolution of device conductances.

A single crossbar array consists of $P \times Q$ nodes (each node representing a synapse), where single node has $2N$ memristors arranged using the differential architecture ($N$ potentiation, $N$ depression devices).
Each memristor state is represented by four variables, $t_p$ for storing the last time the device is written (which is used to calculate the effect of the drift), $count$ for counting how many times it has written (to be used later in the arbiter of N-memristor architectures), $P_{mem}$ for its programming history (required by the \ac{PCM} model) and $G$ for representing the conductance of the the device at $T_0$ seconds later after the last programming time.
The initial conductances of \ac{PCM} devices in the crossbar array are assumed to be iterativelly programmed to \ac{HRS}, sampled from a Normal distribution $\mathcal{N}(\mu=0.1,\sigma=0.01)$ $\mu$S.

The \ac{PCM} crossbar simulation framework supports three major functions: READ, SET and RESET.
The READ function takes the pulse time of the applied READ pulse, $t$, and calculates the effect of drift based on the last programming time $t_p$. 
Then, it adds the conductance-dependent READ noise and returns the conductance values of whole array.
The SET function takes the timing information of the applied SET pulse, together with a mask of shape $(2 \times N \times P \times Q)$ and calculates the effect of the application of a single SET pulse on the \ac{PCM} devices that are selected with the mask.
Finally, the RESET function initializes all the state variables of devices selected with a mask and initializes the conductances using a Normal distribution $\mathcal{N}(\mu=0.1,\sigma=0.01)$ $\mu$S.

\paragraph*{Supplementary Note 2}
\label{section:xbarperf}
READ and WRITE operations to simulated \ac{PCM} devices in the
crossbar model are stochastic and subject to the temporal conductance drift.
Additionally, \ac{PCM} devices offer a very limited bit precision.
Therefore, to ease the network training procedure, especially the hyperparameter
tuning, we developed the \texttt{perf-mode}.
When crossbar model is operated in \texttt{perf-mode}, all stochasticity sources
and the conductance drift are disabled. READ operations directly access
the device conductance without $1/f$ noise and drift, whereas SET
operations increase the device conductance as

\begin{equation}
  G_{N} = G_{N-1} + \frac{G_{MAX}}{2^{CB_{RES}}},
\end{equation}

where, $G_{MAX}$ is the maximum \ac{PCM} conductance set to 12 $\mu$S (conductivity boundaries are determined based on the device measurements  from \cite{Nandakumar_etal18}), and
$CB_{RES}$ is the desired bit-resolution of a single \ac{PCM} device. 
In a nutshell, the \texttt{perf-mode} turns \acp{PCM} into an ideal memory cells corresponding to a digital memory with a limited bit precision.

\begin{figure}[!ht]
  \centering
  \begin{subfigure}{0.5\textwidth}
    \centering
    \includegraphics[width=0.9\linewidth]{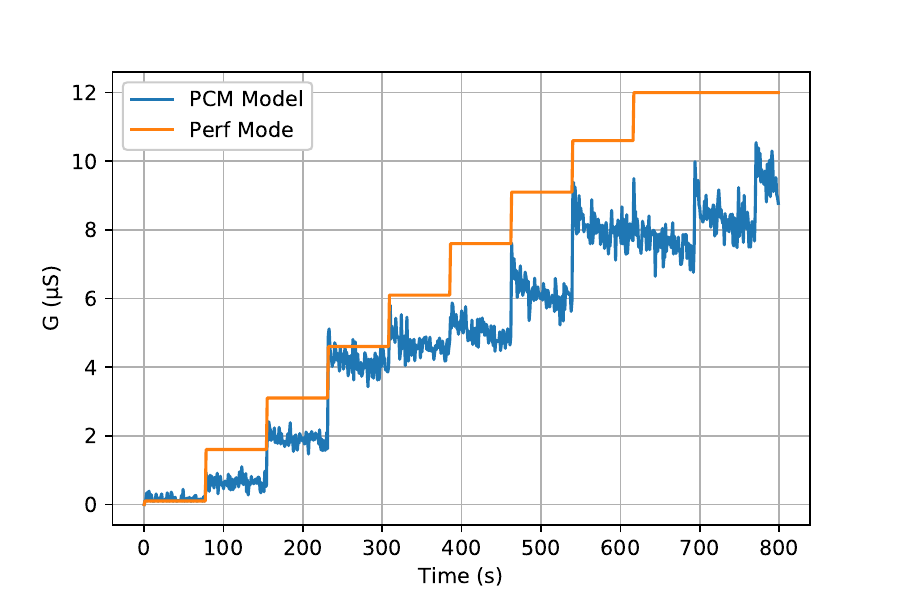}
    \caption{Comparison of the full \ac{PCM} model and its\newline \texttt{perf-mode} equivalent after 8  consecutive SET\newline pulses.}
    \label{fig:sub1}
  \end{subfigure}%
  \begin{subfigure}{0.5\textwidth}
    \centering
    \includegraphics[width=0.9\linewidth]{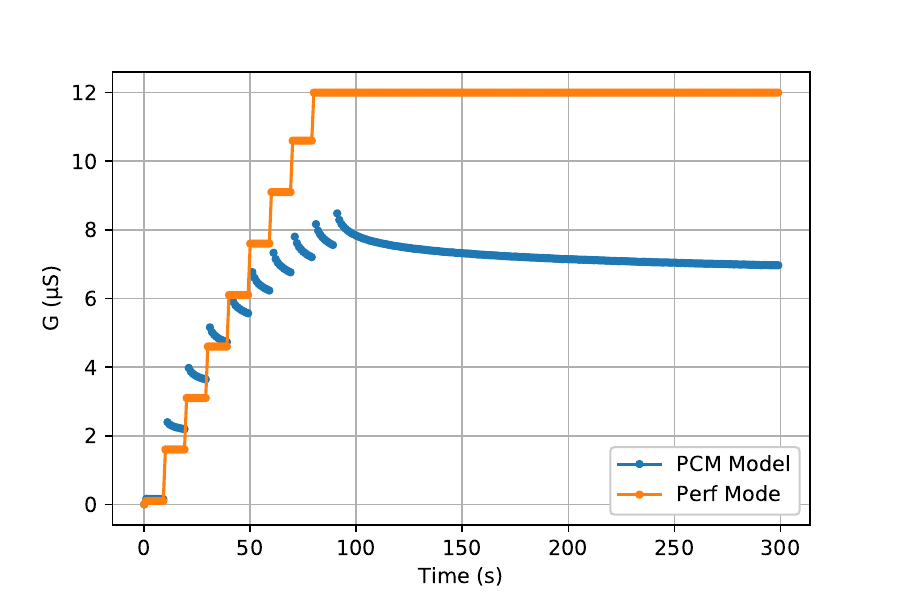}
    \caption{Comparison of the full \ac{PCM} model and its\newline \texttt{perf-mode} equivalent after 8  consecutive SET\newline pulses, averaged over 300 measurements showing \newline the effect of drift.}
    \label{fig:sub2}
  \end{subfigure}
  \caption{The \ac{PCM} crossbar model supports both the full \ac{PCM} model from~\cite{Nandakumar_etal18} and its corresponding simplified version as an ideal digital memory in \texttt{perf-mode}.}
  \label{fig:test}
\end{figure}

\paragraph{Supplementary Note 3}
\label{section:transfer}
Here, we demonstrate the impact of using multiple memristor devices per synapse
(arranged in differential configuration) on the precision of targeted
programming updates.
Specifically, we modeled synapses with $N={1,4,8}$ \ac{PCM} devices and
programmed them from initial conditions of integer conductance values
$G_{source} \in \{-10,10\}$ $\mu$ S  to integer conductance values $G_{target} \in
\{-10,10\}$ $\mu$ S using the multi-memristor update scheme described in Section~\ref{section:updates}. 
The effective conductance of a synapse is calculated by $G_{syn} = \sum_{i=0}^N G_i^+ - \sum_{i=0}^N G_i^-$, however we normalized the
conductance across 1-PCM, 4-PCM and 8-PCM architectures for an easier comparison,
such that $G_{syn} = \frac{1}{N}(\sum_{i=0}^N G_i^+ - \sum_{i=0}^N G_i^-)$.

Our empirical results verifies the claim of Boybat et al. \cite{Boybat_etal18} that the standard deviation and the update resolution of the write process decreases by $\sqrt{N}$.  

\label{section:wtransfer}
\begin{figure}
  
  \includegraphics[width=1\textwidth]{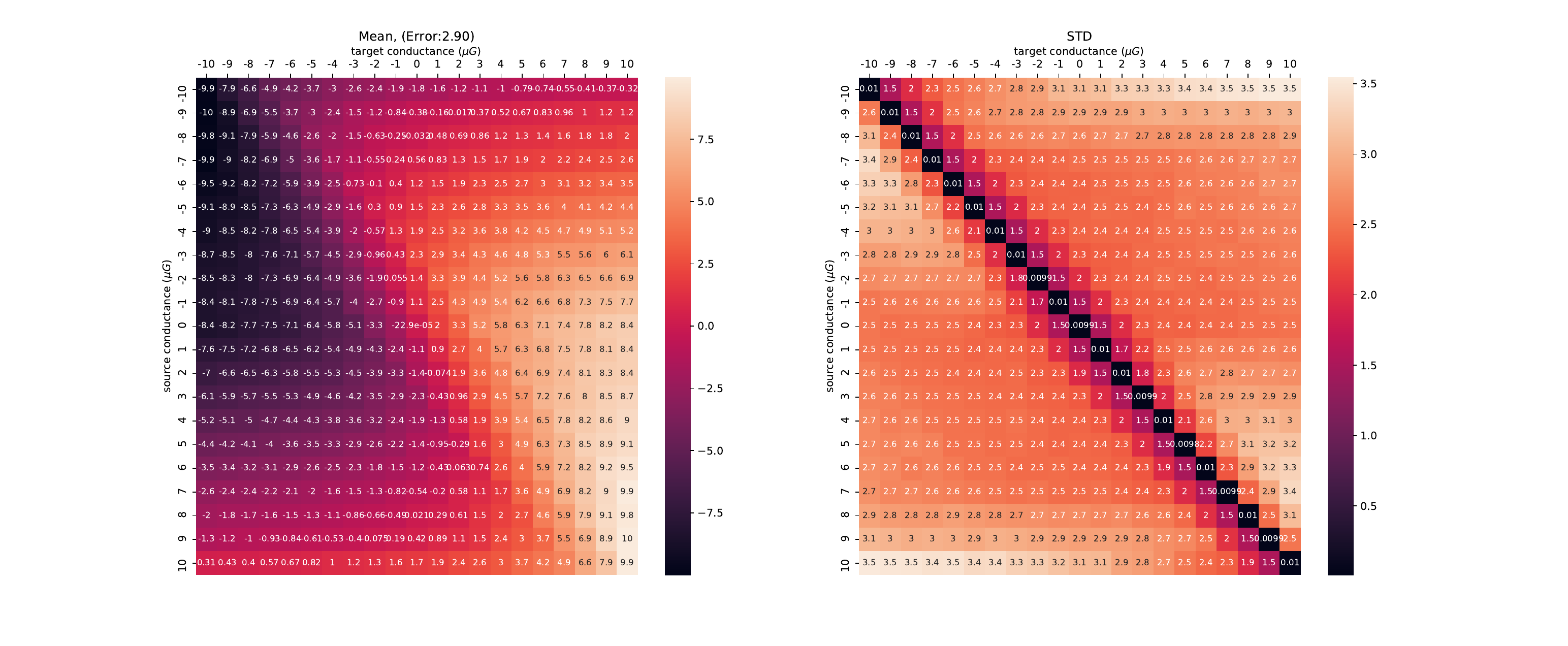}
  \caption{Multi-memristor configuration with 1 \ac{PCM} (one depression and one
    potentiation) per synapse}
  \label{fig:multi1}
\end{figure}

\begin{figure}
  
  \includegraphics[width=1\textwidth]{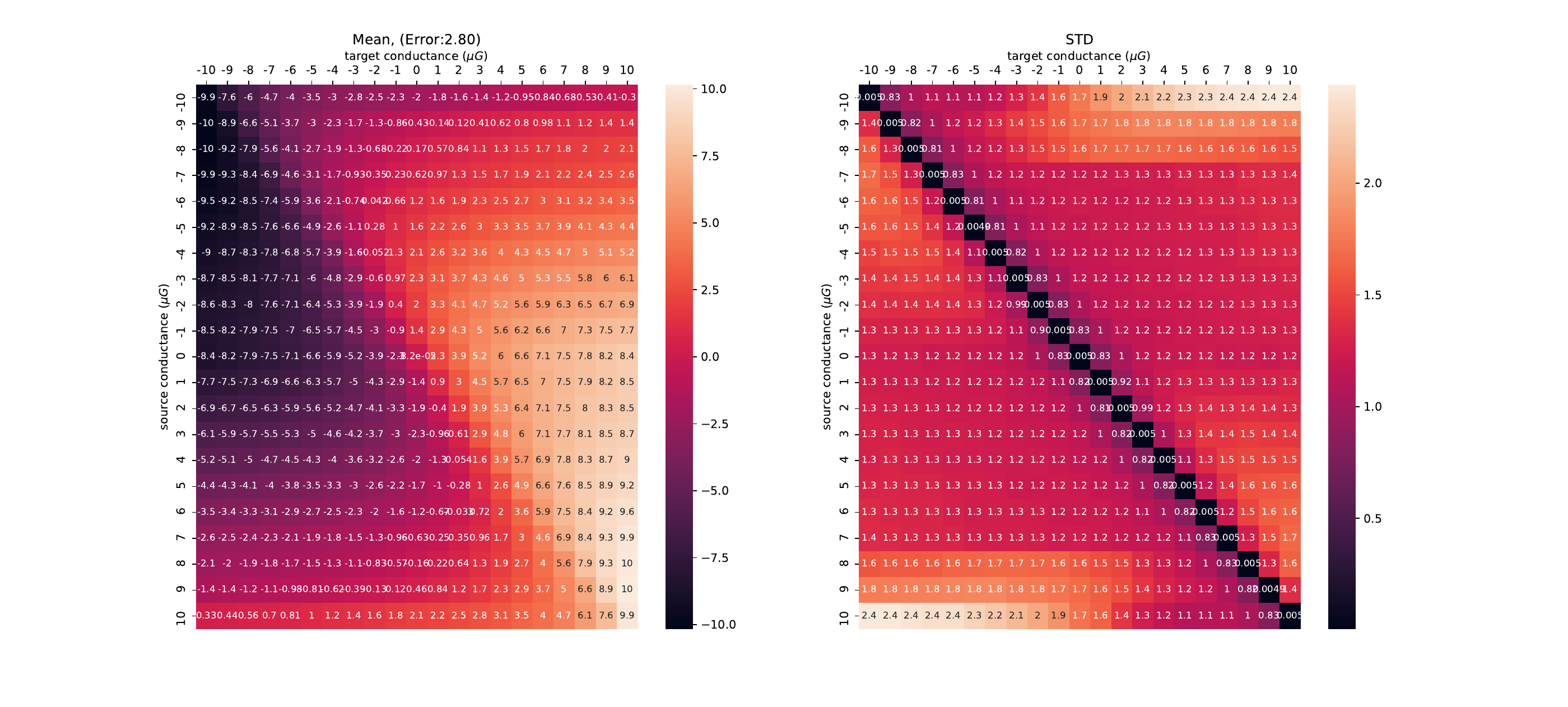}
  \caption{Multi-memristor configuration with 8 \ac{PCM} (four depression and four
    potentiation) per synapse}
  \label{fig:multi4}
\end{figure}

\begin{figure}
  
  \includegraphics[width=1\textwidth]{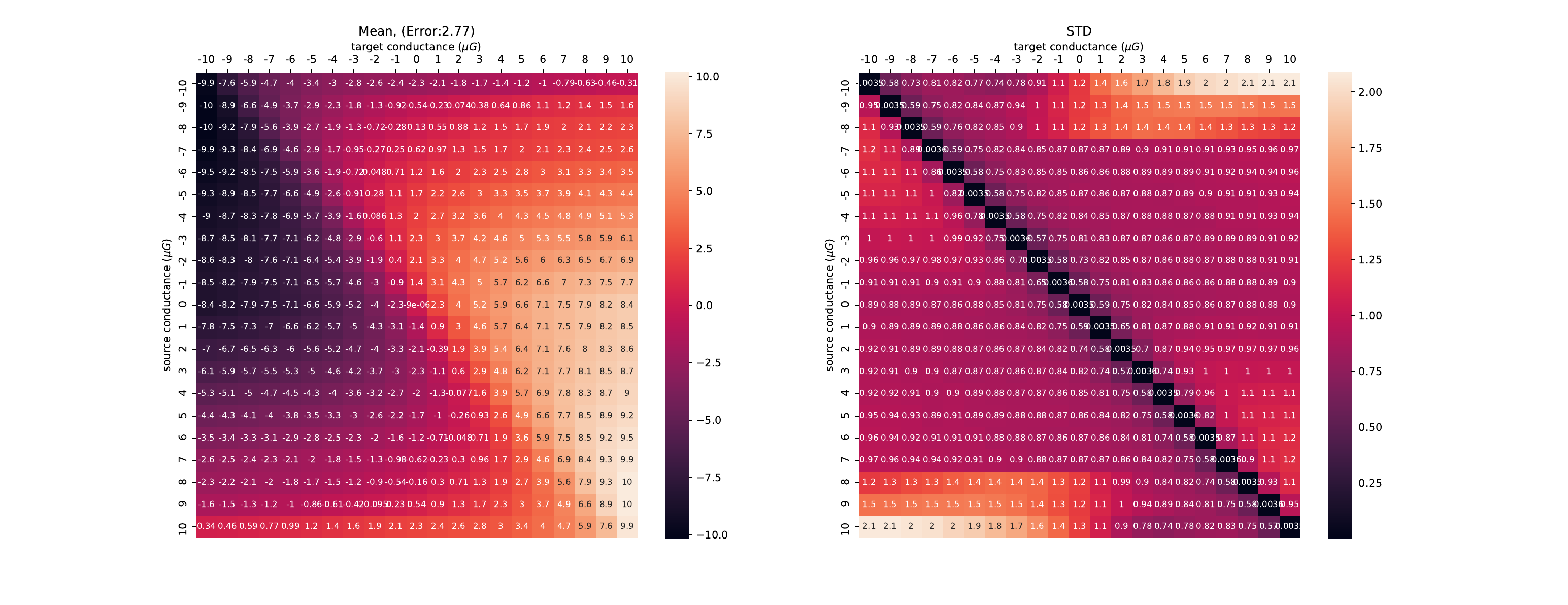}
  \caption{Multi-memristor configuration with 16 \ac{PCM} (eight depression and eight
    potentiation) per synapse}
  \label{fig:multi8}
\end{figure}

\paragraph{Supplementary Note 4}
\label{section:upd-ready}
In the differential architectures, consecutive SET pulses applied to positive and negative memristors may cause the saturation of the synaptic conductance and block further updates.
The saturation effect is more apparent when a single synapse gets 10+ updates in one direction (potentiation or depression) during the training.
For example, this effect is clearly visible in SupplFig.~\ref{fig:multi1}, Fig.~\ref{fig:multi4} and Fig.~\ref{fig:multi8}, when the source conductance and target conductances differ by more than 8-10 $\mu$S.

We implemented a weight update scheme denoted as the update-ready criterion, which aims to prevent conductance saturation while applying single large updates.
Before doing the update, we read both positive and negative pair conductances, and check if the target update is possible. 
If not, we reset both devices, calculate the new target and apply the number of pulses accordingly.
For example, given $G^+=8 \mu$S and $G^-=4 \mu$S and the targeted update $+6 \mu$S, the algorithm decides to reset both devices because $G^+$ can't be increased up to $14 \mu$S. After both devices are reset, $G^+$ can be programmed to $10 \mu$S).
Although our \ac{PCM} crossbar array simulation framework supports it, this weight transfer criterion is not used in our simulations because it requires reading the device states during the update.

\begin{figure}[H]
  
  \includegraphics[width=1\textwidth]{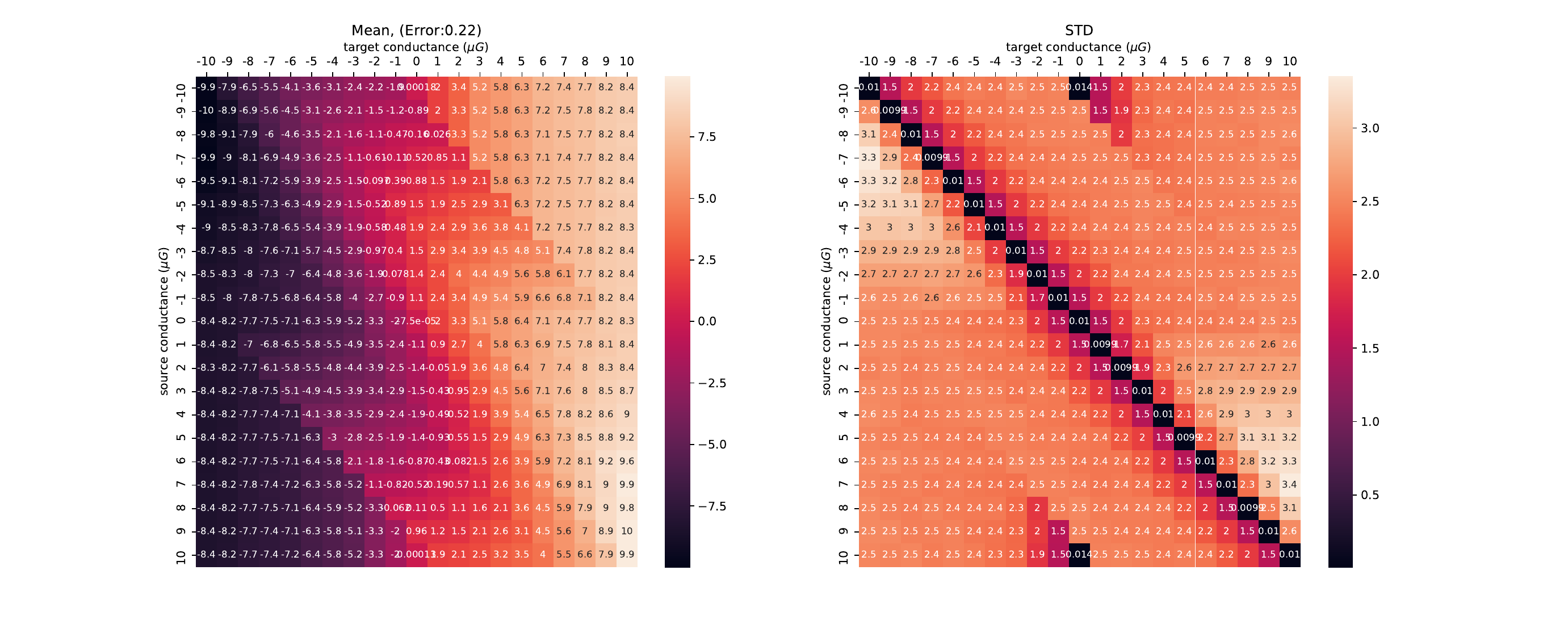}
  \caption{Update-ready criterion tested with $N=1$ memristor per
    synapse.}
  \label{fig:updready}
\end{figure}

\paragraph{Supplementary Note 5}
\label{section:loss-eval}

We have defined the task success criteria as MSE Loss $<0.1$ based on visual inspection. Below in Fig~\ref{fig:visualloss}, some network performances are shown.

\begin{figure}[H]
\centering
\begin{subfigure}{0.5\textwidth}
    \centering
    \includegraphics[width=1\textwidth]{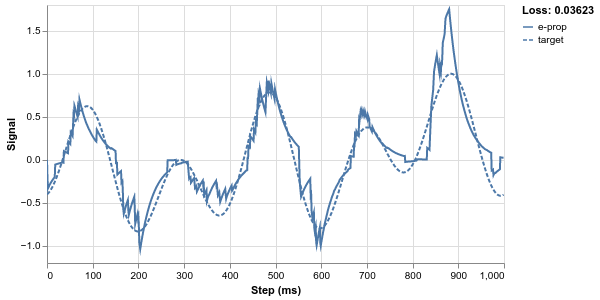}
\end{subfigure}%
\begin{subfigure}{0.5\textwidth}
    \centering
    \includegraphics[width=1\textwidth]{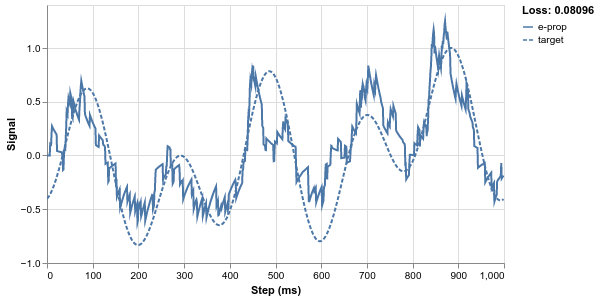}
\end{subfigure}
\begin{subfigure}{.5\textwidth}
    \centering
    \includegraphics[width=1\textwidth]{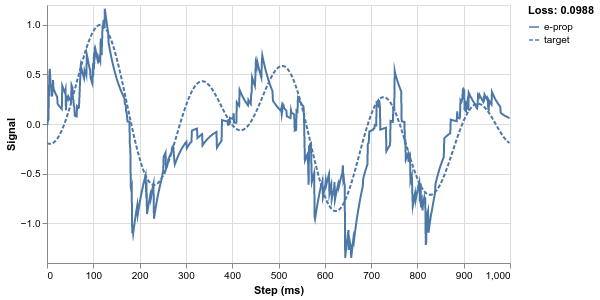}
\end{subfigure}%
\begin{subfigure}{.5\textwidth}
    \centering
    \includegraphics[width=1\textwidth]{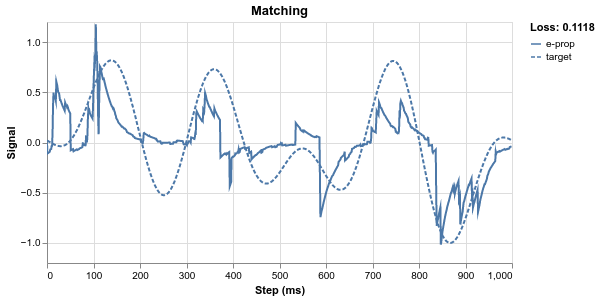}
\end{subfigure}
\begin{subfigure}{.5\textwidth}
    \centering
    \includegraphics[width=1\textwidth]{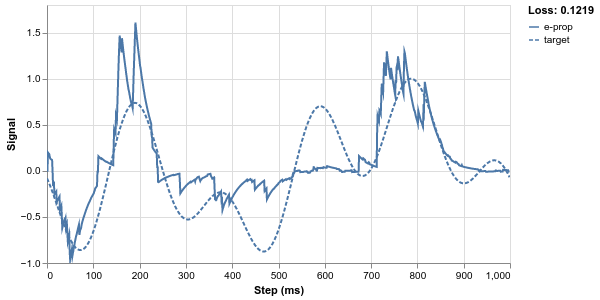}
\end{subfigure}%
\begin{subfigure}{.5\textwidth}
    \centering
    \includegraphics[width=1\textwidth]{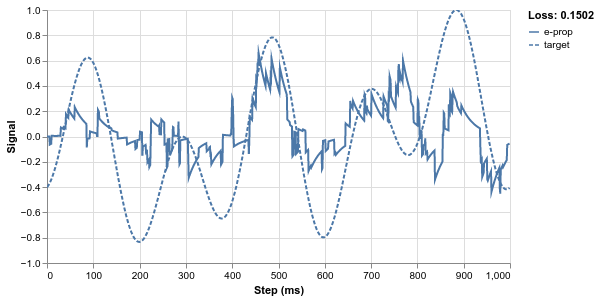}
\end{subfigure}
\caption[short]{Comparison of network performances with six different loss values.}
\label{fig:visualloss}
\end{figure}

\paragraph{Supplementary Note 6}
\label{section:networkstatsx}

\begin{figure}[H]
  \centering
  \includegraphics[width=0.6\textwidth]{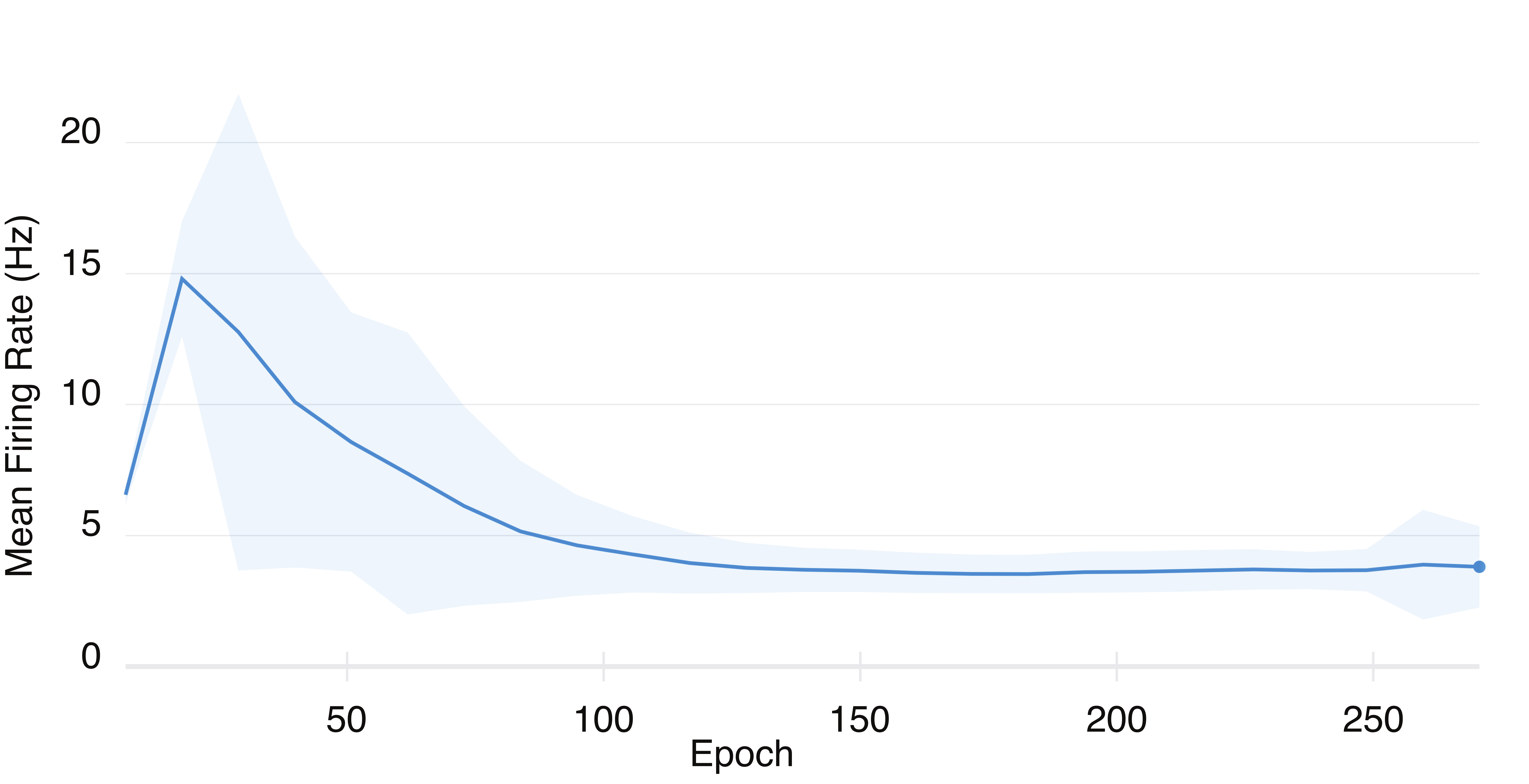}
  \caption{Mean firing rate of 50 networks with \ac{PCM} synapses trained using the mixed-precision method.}
  \label{fig:fr}
\end{figure}

\begin{figure}[H]
  \includegraphics[width=0.6\textwidth]{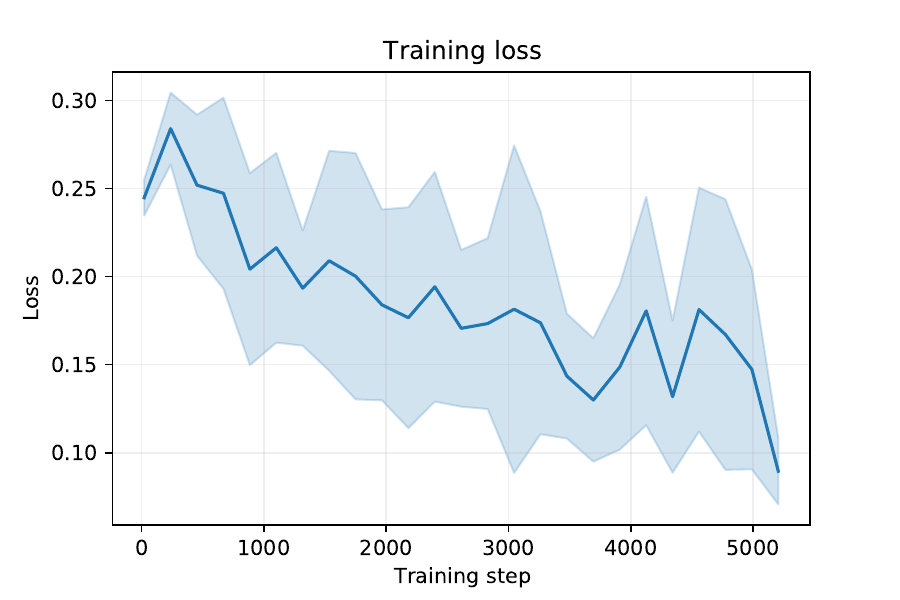}
  \caption{MSE loss of 50 networks trained with \ac{PCM} synapses using the mixed-precision method.}
  \label{fig:losstr}
\end{figure}

\newpage
\section*{Appendix 2: Mosaic: in-memory computing and routing for small-world spike-based neuromorphic systems}

\paragraph{Supplementary Note 1}

\begin{figure}[H]
\includegraphics[width=1.0\textwidth]{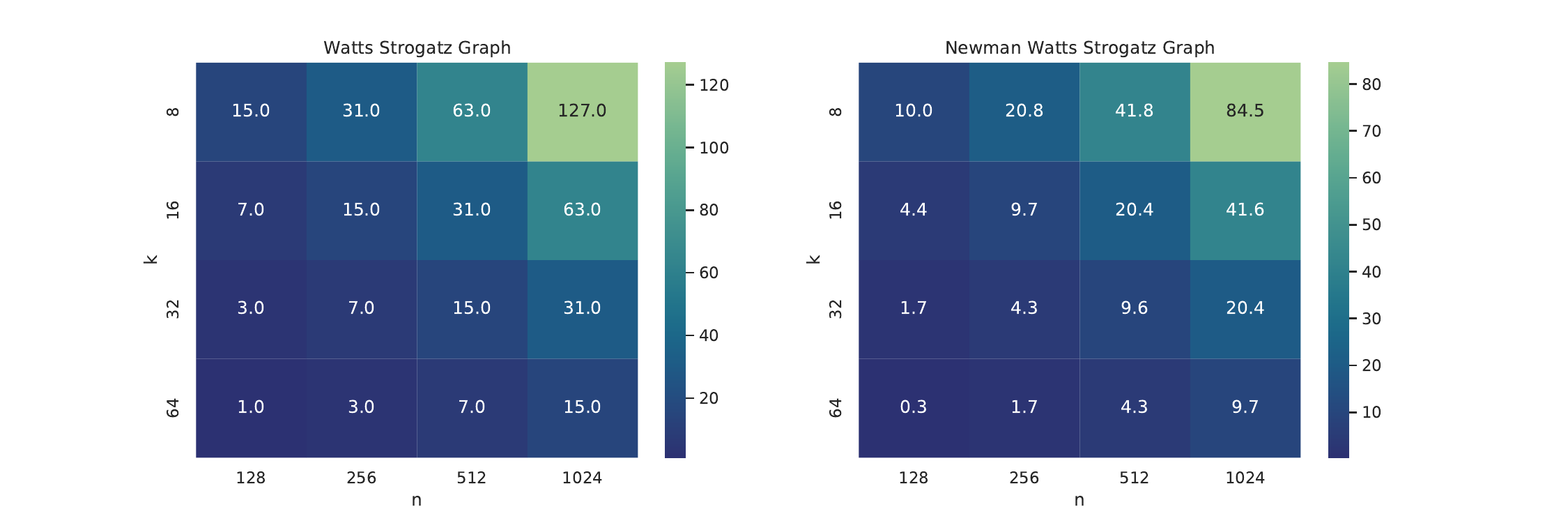}
\centering
\caption{The heatmaps show the ratio of zero elements to non-zero elements in the connectivity matrix for two examples of recurrently connected small-world graph generators. As $n$ (number of nodes, e.g., neurons, in the graph) increases and $k$ (number of neighbour nodes for each node in a ring topology) decreases, the more connections in the connectivity matrix will be zero, indicating the increased proportion of non-used memory elements in a $n \times n$ crossbar array.}
\label{supfig:ratio_empty}
\end{figure}

Figure \ref{supfig:ratio_empty} quantifies the under-utilization of conventional crossbar arrays while storing example small-world connectivity patterns generated by two standard random graph generation models: Watts-Strogatz small-world  graphs~\cite{watts_strogatz1998_smallworld} and Newman-Watts-Strogatz small-world graphs~\cite{Newman1999}.
The first type of graphs is characterized by a high degree of local clustering with short vertex–vertex distances, observed in neural networks and self-organizing systems, whereas the latter type mostly captures the properties of lattices associated with statistical physics.

\paragraph{Supplementary Note 2}

To communicate the events between the computing nodes in neuromorphic chips, \ac{AER} communication scheme has been developed and used~\cite{Boahen_etal98}. 
In \ac{AER}, whenever a spiking neuron in a chip (or module) generates a spike, its ``address'' (or any given ID) is written on a high-speed digital bus and sent to the receiving neuron(s) in one (or more) receiver module(s). In general, \ac{AER} processing modules require at least one \ac{AER} input port and one \ac{AER} output port. As neuromorphic systems scale up in size, complexity, and functionality, researchers have been developing more complex and smarter \ac{AER} ``variations'' to maintain the efficiency, reconfigurability, and reliability of the ever-growing target systems they want to build.
The scheme that is used to transport events can be source or destination based, where the source or destination address is embedded in the sent event ``packet''. In the source-based scheme, each receiving neuron has a local \ac{CAM} that stores the address of all the neurons that are connected to it. In the destination-based approach, each event hops between the nodes where its address gets compared to the node's address until it matches and thus gets delivered.
Source-driven routing provides the designer with more freedom to balance event traffic and design routes, but the hardware complexity increases the delays.
Destination-based creates pre-determined routes along the network and the designer can only change the output ports~\cite{Zamarreno-Ramos_etal11}. 
In summary, in source-based routing, the system requires a \ac{CAM} memory per neuron, which results in an increase in the area and memory access read times. In destination-based routing, the configurability in the network structure is reduced. Comparatively, in the Mosaic, the routers are memory crossbars that are distributed between the computing cores and steer the spiking information in the mesh. Thus, neither local CAMs, nor a centralized memory is required for routing.

\paragraph{Supplementary Note 3}

\begin{figure}[H]
  \centering
  \includegraphics[width=0.35\textwidth]{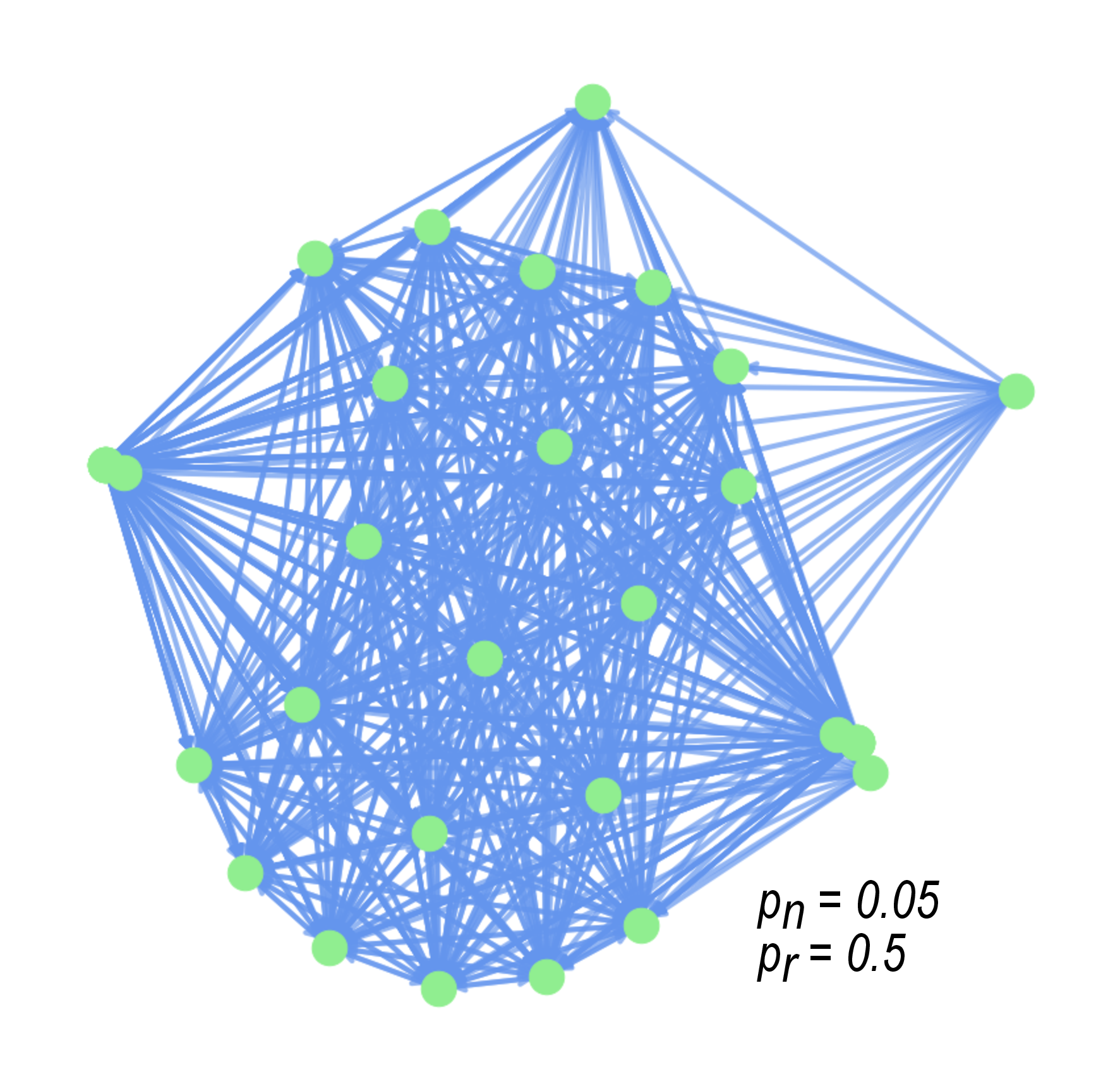}
  \includegraphics[width=0.35\textwidth]{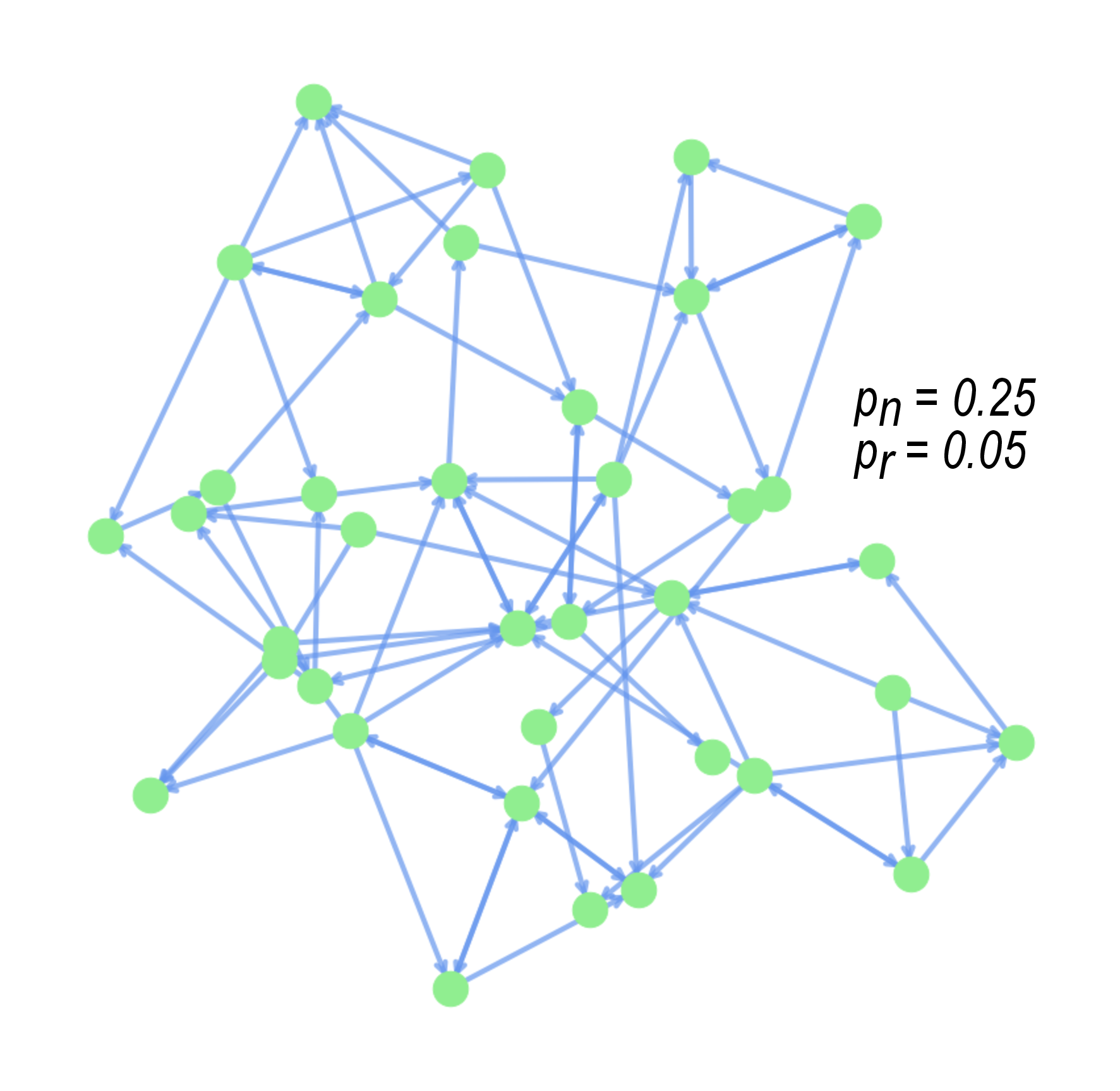}
  \includegraphics[width=0.7\textwidth]{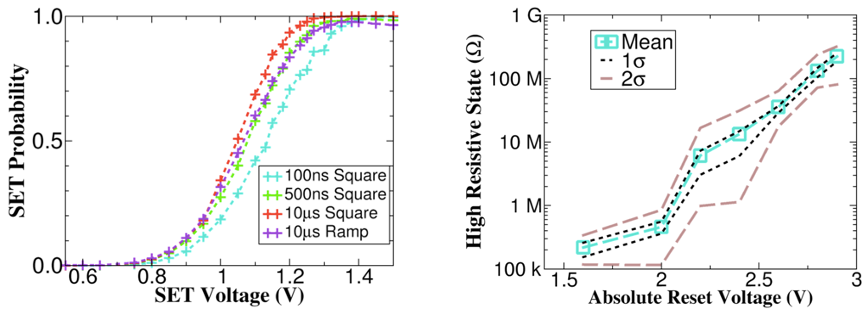}

  \caption{(top) Different random graphs generated using Mosaic model, changing the probability of devices being in their High Conductive state in the neuron tile ($p_n$) and routing tile ($p_r$). (bottom) The probability of device switching is a function of the voltage applied to it while being programmed.}
  \label{supfig:graphs}
\end{figure}

\begin{figure}[H]
    \centering
    \includegraphics[width=0.7\linewidth]{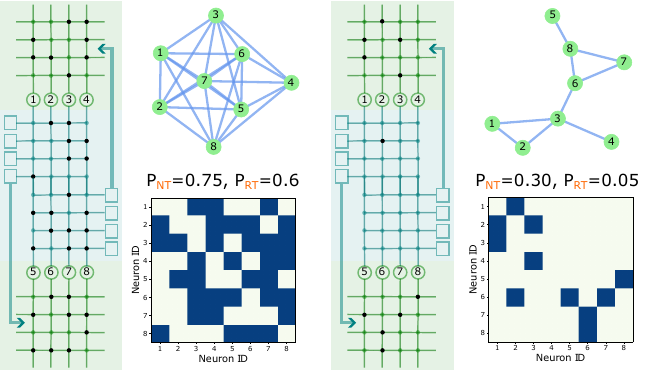}
    \caption{Mosaic connectivity example, formed by setting the probability of connection within Neuron Tile ($p_{NT}$) and Routing Tiles ($p_{RT}$). (left) Densely connected Mosaic composed of 2 Neuron Tiles and 1 Routing Tile. The graph related to its connectivity is shown as well adjacency matrix. (right) Sparsely connected Mosaic. The graph is programmed to favor the intra-Neuton Tile connectivity and allow for two clusters to emerge, penalizing connections between the two clusters.}
    \label{suppfig:Mosaic_graph}
 \end{figure}

Routing tiles define the connectivity of spiking neural networks implemented on Mosaic. When the number of memristive devices in the routing tiles which are in their high-conductive state (HCS) is not sparse, Mosaic resembles a densely connected neural network (Fig.~\ref{supfig:graphs}, top left). When most of the memristor in the routing tiles are in the low-conductance state, Mosaic is sparsely connected (Fig.~\ref{supfig:graphs}, top right). Furthermore, one can further sparsify Mosaic networks by setting memristors in the neuron tiles to the LCS. 
To do so, we can change the probability of memristors being in their HCS in the neuron tiles, $p_n$, and in the routing tiles,  $p_r$. The switching of the \acp{RRAM} presents the property of probabilistic switching as a function of the voltage applied during the programming operation as is shown in Fig.~\ref{supfig:graphs}, bottom.

Fig.~\ref{suppfig:Mosaic_graph} shows the construction of two graph topologies, made of 2 Neuron Tiles and one Routing Tile, to clarify the formation of the graphical structure in the Mosaic. By controlling the probability of connections within the Neuron and Routing Tiles, we can produce a densely connected graph (left) with $p_{NT}=0.75$, $p_{RT}=0.6$, and a sparse graph (right) with $p_{NT}=0.30$, $p_{RT}=0.05$. 

The corresponding connectivity matrix is also shown in the figure, which is directly represented as a hardware architecture in the 3 tiles of the Mosaic, as shown in the figure.

\paragraph{Supplementary Note 4}

\begin{figure}[H]
  \centering
  \includegraphics[width=0.9\textwidth]{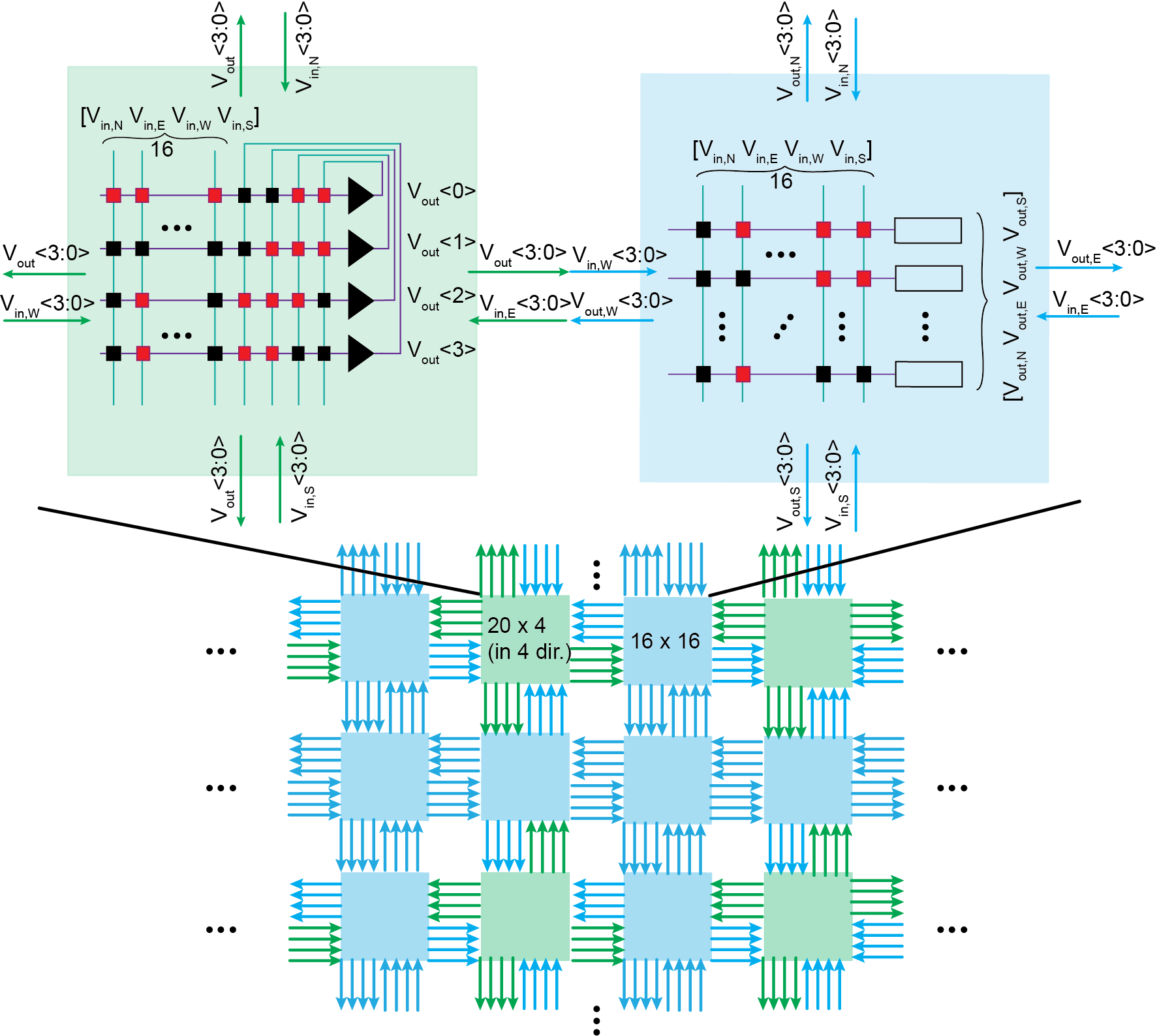}
  \caption{(Neuron tiles (green) transfer information in the form of spikes to each other through routing tiles (blue). 
  %(top) Each tile can send and receive information from four directions (North (N), South (S), West (W) and East (E). 
 Details of the Mosaic architecture is shown with the size of the neuron and routing tiles. The neuron tiles receive feed-forward input from four directions of North (N), East (E), West (W), and South (S), and local recurrent input from the neurons in the tile. The neurons integrate the information and once spike, send their output to 4 directions. Having 4 neurons in a tile, gives rise to 16 outputs (4 outputs copied in 4 directions), and 20 inputs (4 inputs from 4 directions (16), plus 4 recurrent inputs). The routing tiles receive 16 inputs (4 inputs from 4 directions) and send out 16 outputs (4 outputs in 4 directions). In the crossbars, the red squares and black squares represent devices in their high conductive and low conductive state, respectively. The connection between the neuron tile and the routing tile is directly through a wire. For instance, $V_{\text{out}}\!<\!3\text{:}0\!>$ is the same as the $V_{in,W}$, and $V_{\text{in,E}}\!<\!3\text{:}0\!>$ is the same as $V_{out,W}$.  }
  \label{supfig:nwse}
\end{figure}

Figure \ref{supfig:nwse} shows the details of the Mosaic architecture, with a zoomed in neuron and routing tile pair. The diagram in the top shows how one cluster of neuron/one router sends and receives information to and from the routing/neuron tile. This highlights the strength of this architecture which makes the connectivity easy through simple wiring to the neighbour, without suffering from long wires, as the maximum length of a wire is the size of the wire from one row/column, plus the size of the connecting column/row. 

\paragraph{Supplementary Note 5} 

\begin{figure}[H]
\includegraphics[width=1\textwidth]{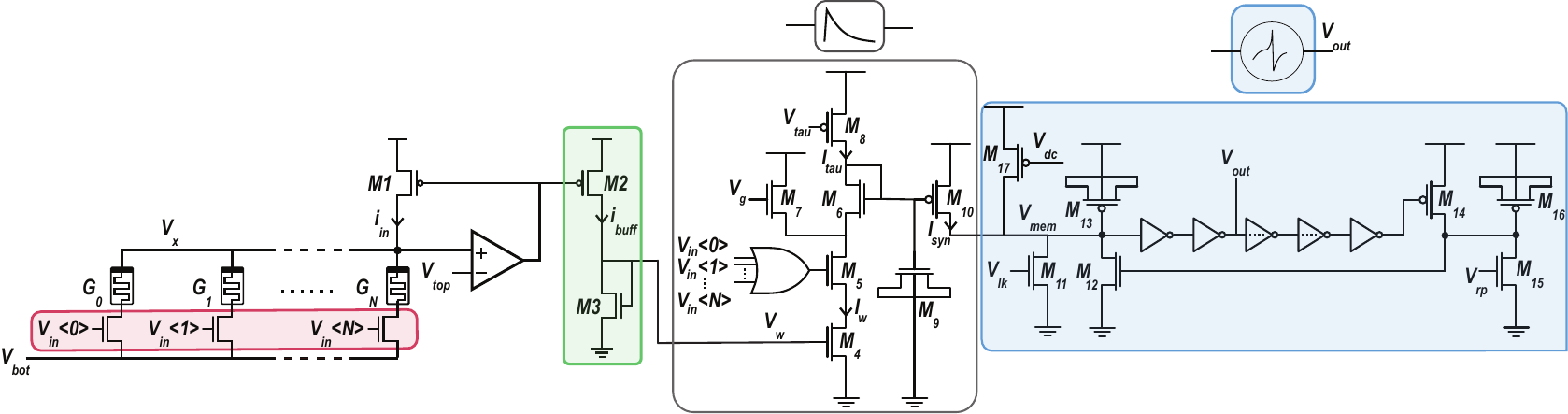}
\centering
  
\caption{ Schematic of the neuron tile including the CMOS synapse and neuron circuits fabricated for use in this paper. RRAMs are used as the weights of the neurons. On the arrival of any of the input events $V_{in<i>}$, the amplifier pins node $V_x$ to $V_{top}$ and thus a read voltage equivalent to  $V_{top}-V_{bot}$ is applied across $G_i$, giving rise to current $i_{in}$ at $M_1$. This current is mirrored to $M_2$ giving rise to $i_{buff}$ which is in turn again mirrored through the $M_3-M_4$ transistor pair. 
The ``synaptic dynamics'' circuit is the Differential Pair Integrator (DPI)~\cite{Bartolozzi_Indiveri2007}. On the arrival of any of the input events, $V_{i}, 0<i<n$, $I_w$, equivalent to $i_{buff}$, flows in transistor $M5$. Depending on the value on $V_g$, a portion of $I_w$ flows out of the MOS capacitor M6 and discharges it. This current is proportional to $G_i, 0<i<n$. As soon as the event is gone, MOS capacitor $M_6$ charges back through the $M_8$ path with current $I_{tau}$, which determines the rate of charging, and thus the time constant of the synaptic dynamics. The output current of the DPI synapse, $I_{syn}$, is injected into the neuron's membrane potential node, $V_{mem}$, and charges MOS capacitor $M_{13}$. There is also an alternative path with a DC current input through $M_{17}$ which can charge neuron's membrane potential. Membrane potential charging has a time constant determined by $V_{lk}$ at the gate of $M_{11}$. As soon as the voltage developed on $V_{mem}$ passes the threshold of the following inverter stage, it generates a pulse. The width of the pulse, depends on the delay of the feedback path from $V_{out}$ to the gate of $M_{12}$. This delay is determined by the inverter delays, and the refractory time constant. The inverter symbols with the horizontal dashed lines correspond to a starved inverter circuits with longer delays. The refractory period time constant depends on the MOS cap $M_{16}$ and the bias on $V_{rp}$. 
}
\label{supfig:neuron_all}
\end{figure}

\begin{figure}[H]
\includegraphics[width=0.5\textwidth]{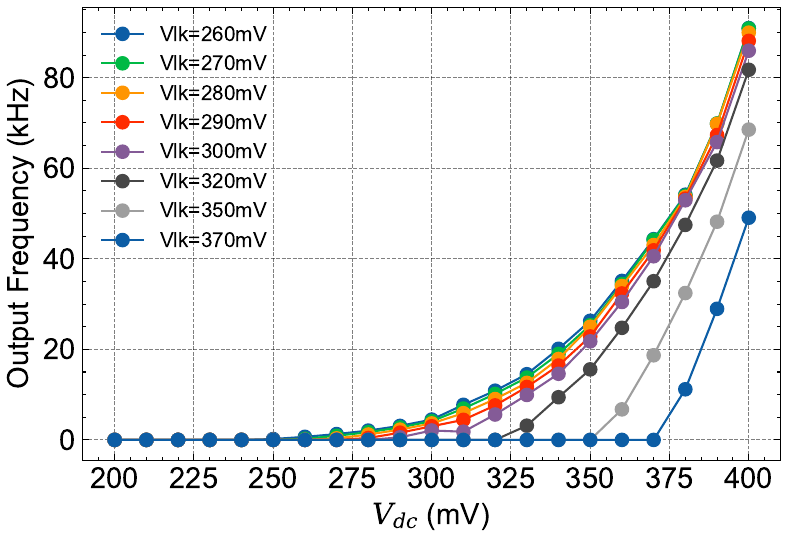}
\centering
  
\caption{Measurements from fabricated neuron's output frequency as a function of the input DC voltage. The DC voltage is applied at the gate of transistor $M_{17}$ shown in Fig.~\ref{supfig:neuron_all} as $V_{dc}$. Therefore, as the gate voltage of $M_{17}$ changes linearly, the current of $M_{17}$ and thus the output frequency of the neuron changes non-linearly. Each curve is measured with a different neuron's time constant, determined by a different voltage, $V_{lk}$, on the gate of transistor $M_{11}$ in Fig.~\ref{supfig:neuron_all}. As the leak voltage increases, the neuron's time constant decreases, giving rise to a lower output frequency. 
}
\label{supfig:neuron_freq}
\end{figure}

Details of the implementation of the neuron row, the circuit that leverages the information of the conductance of a memristor to weight the effect of a spike to a neuron is shown in Figure \ref{supfig:neuron_all}. The circuit features multiple inputs connected to a row of memristive devices (left) and a Front-End circuit buffering the current read from the devices to a differential-pair-integrator synapse. The synapse is then connected to a leaky-integrated-and-fire (LIF) neuron which eventually emits a spike. Figure \ref{supfig:neuron_freq} delves deeper in the behavior of the LIF neuron analyzing its output spiking frequency against an input DC voltage and its linear behavior respect to the RRAM conductance in a neuron row circuit.

\paragraph{Supplementary Note 6} 

\begin{figure}[H]
  \centering
  \includegraphics[width=0.8\textwidth]{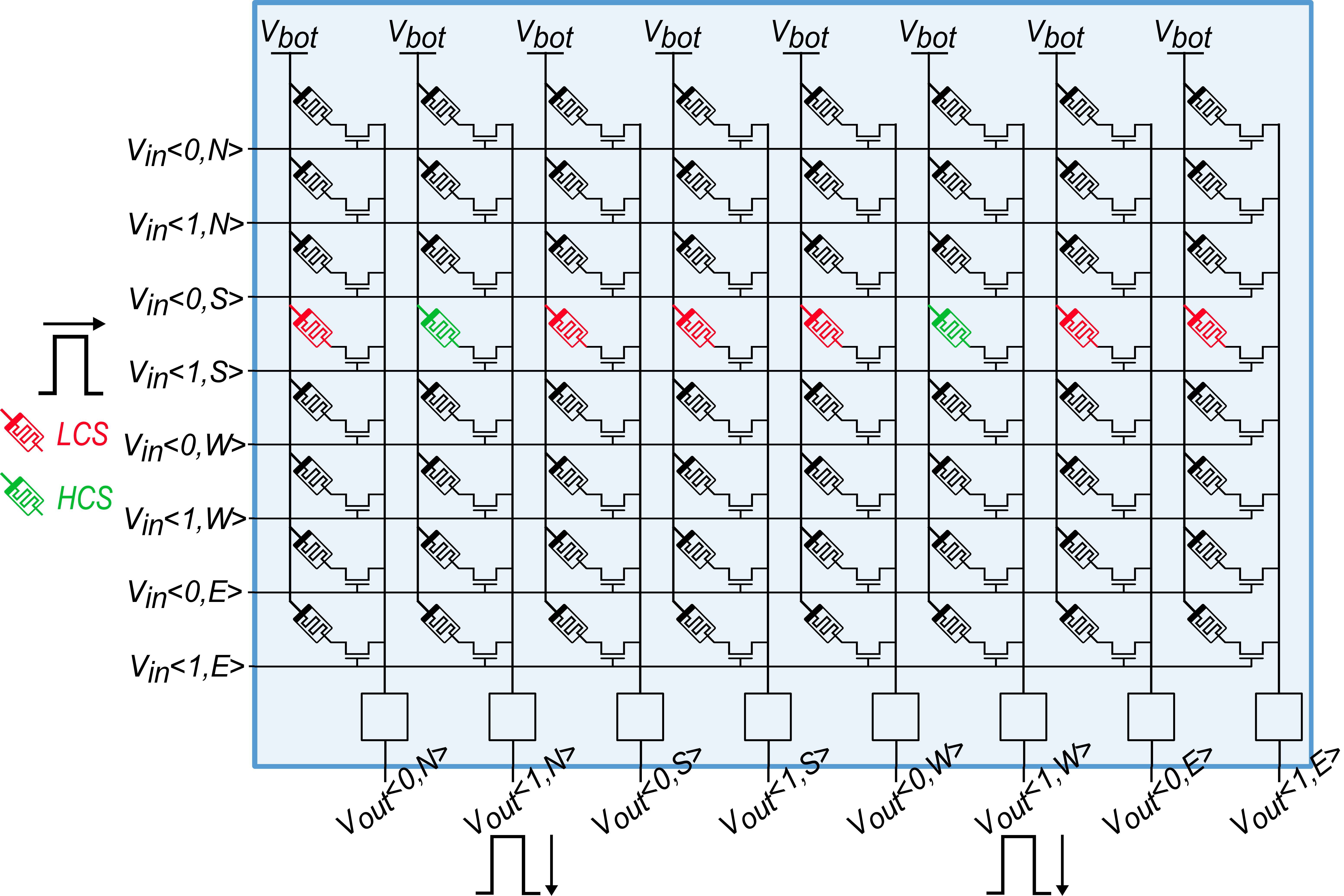}
  \caption{Schematic of a routing tile circuit offering two paths per direction. The routing tile receives eight inputs, comprising two pulse channels per direction, labelled as < 0 > or < 1 >, from the neighbouring tiles to the North (N), South (S), East (E) and West (W), and provides complimentary outputs. An example is shown of an input pulse arriving to the common gate of the fourth row of memory. Devices are coloured green or red to denote whether they are in the \ac{HCS} or \ac{LCS}. It is shown that, due to this input pulse, output pulses are produced by the routing columns containing the (green) devices programmed in the \ac{HCS}. }
  \label{supfig:routing_tile}
\end{figure}

\begin{figure}[H]
  \centering
  \includegraphics[width=0.7\textwidth]{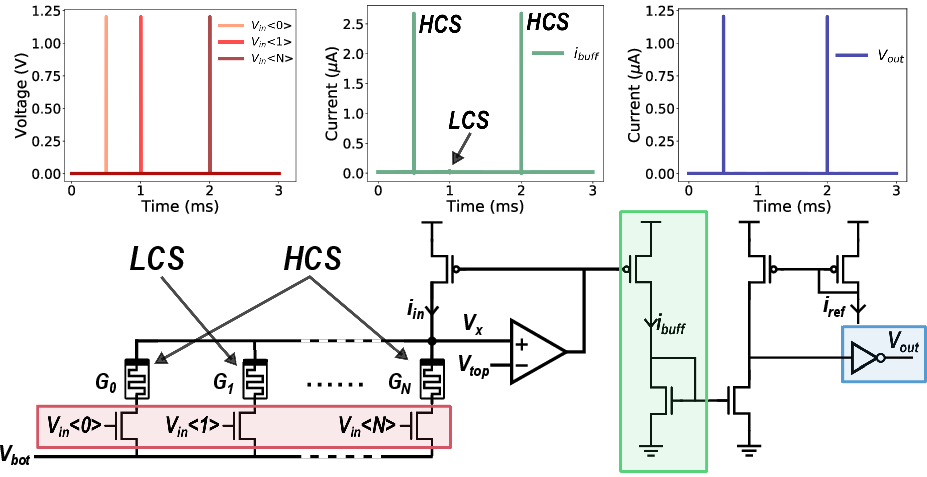}
  \caption{The routing column circuit with example waveforms. Input (red, left) voltage pulses, Vin, draw a current iin proportional to the conductance state, Gn, of the read 1T1R structures. Two devices are labelled with \ac{HCS}, indicating that they have been programmed with a conductance corresponding to the high conductance state, and one is labelled \ac{LCS} in reference to the low conductance state. This resulting currents are buffered (green, centre), ibuff , into a current comparator circuit where it is compared with a reference current iref . When the buffered current exceeds the reference current a voltage pulse is generated at the column output (blue, right).}
  \label{supfig:routing_col}
\end{figure}

Details on the implementation of the Routing Tiles. Figure \ref{supfig:routing_tile} shows a full-size schematics of a routing tile with 2 neurons allocated per direction. Figure \ref{supfig:routing_col} expands on the details of the implementation of the routing column, the circuit that uses the state of a memristor to decide whether to block or pass (route) a spike through the Mosaic architecture.

\newpage
\paragraph{Supplementary Note 7}

\begin{figure}[H]
  \centering
  \includegraphics[width=0.3\textwidth]{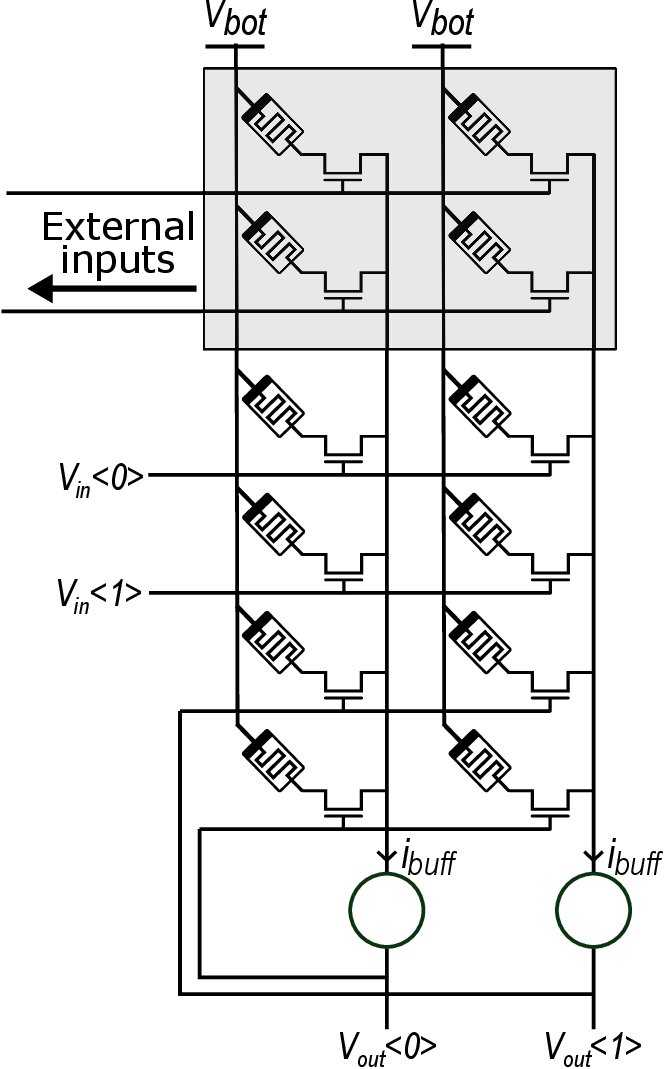}
  \caption{An example of how a neuron tile can be interfaced to external event-based inputs (i.e., those generated by an event-based sensor). With respect to the neuron tile circuit presented in the paper (permitting connections to adjacent tiles as well as recurrent connections within the tile), this figure shows two additional rows of devices stacked on top of the array. As an arbitrary example, here two additional signals can be integrated in the neuron circuits.}
  \label{supfig:extinput_tile}
\end{figure}

\paragraph{Supplementary Note 8}

\begin{figure}[H]
  \centering
  \includegraphics[width=0.85\textwidth]{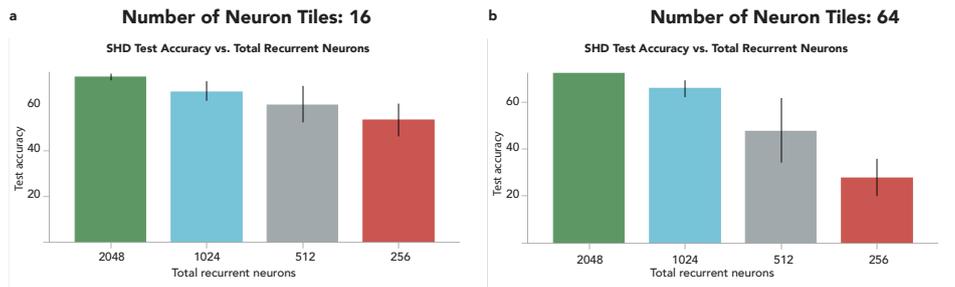}
  \caption{SHD keyword spotting dataset test accuracies for Mosaic architectures with different total number of neurons in the network for a) 4x4 Neuron Tile layout (a total of 16 number of Neuron tiles) and b) 8x8 Neuron Tile layout. The number of neurons per tile is equal to the total number of recurrent neurons divided by the number of neuron tiles. Median and standard deviation are calculated using 3 experiments with varying sparsity constraints.}
  \label{supfig:tiles-bm}
\end{figure}

\begin{figure}[H]
  \centering
  \includegraphics[width=0.85\textwidth]{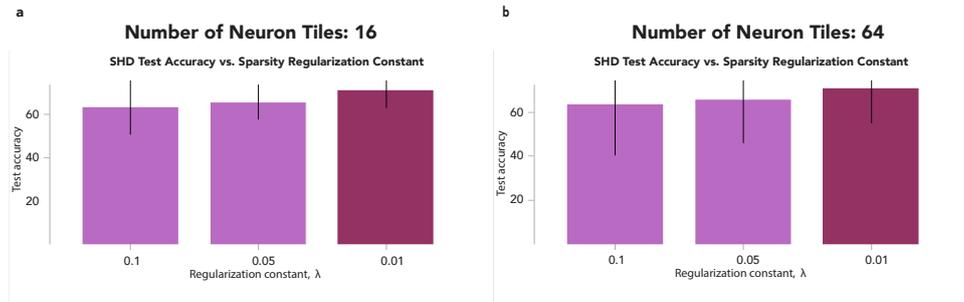}
  \caption{SHD keyword spotting dataset test accuracies for Mosaic architectures trained with different sparsity regularization values. As explained in the Methods section of the main text, the regularization is added to the loss function to exponentially penalize the long-range connections.  The plot shows the accuracy for strong (default, $\lambda=0.1$), medium ($\lambda=0.05$) and weaker ($\lambda=0.01$) sparsity regularization on a) 4x4 neuron tile layout and b) 8x8 neuron tile layout. Median and standard deviation are calculated using 4 experiments with varying number of neurons per neuron tile.}
  \label{supfig:sparsity-bm}
\end{figure}

\paragraph{Supplementary Note 9}

\begin{figure}[H]
  \centering
  \includegraphics[width=0.5\textwidth]{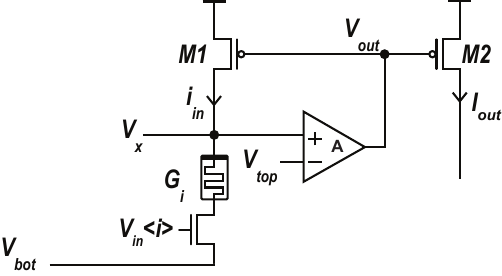}
  \caption{Analysis of the read-out circuitry. The amplifier with gain A, pins voltage $V_x$ to the voltage $V_{top}$. On the arrival of the pulse on $V_{in}<i>$, a current equal to $i_{in}=(V_{top}-V_{bot})G_i$ flows into the memristor $i$, which is then mirrored out $i_{out}$.}
  \label{supfig:readout}
\end{figure}

Figure~\ref{supfig:readout} details the implementation of the read-out circuit used in the Mosaic architecture. Though not optimized for area, we have used this implementation for both the neuron and routing tiles. 

The dominant power consumption of the circuit depends on the required bandwidth (BW) of the feedback loop. This BW depends on the maximum conductance of the RRAM, $G_{max}$. 
For $G_{i,{max}}$, once an input arrives to $V_{in}<i>$, the current $i_{in}$ has to settle to $({V_{top}-V_{bot})} G_{i,max}$ within a settling time, $t_s$, a proportion of the pulse width. This timing sets the speed at which the loop should work, and thus its BW. If the loop does not close in this time, the amplifier will slew, and the voltage $V_x$ drops. 
In both neuron and routing tiles, this condition should be met for $V_x$ to stay pinned at $V_{top}$ while RRAM is being read. However, the neuron and routing tiles have different BW requirements.

In the \textbf{neuron tile}, the read-out circuitry has to resolve between at least 8 levels of current for the 8 levels that each RRAM device can take. Therefore, the Least Significant Bit (LSB) of the $i_{in}$ current for the neuron tile is $i_{in,{LSB},N} = \frac{V_{ref} (G_{max}-G_{min})}{N}$. Based on the Fig.~\ref{fig:exp_neuron}d, this value for the neuron tile is $\frac{100 mV (120 \mu S - 40 \mu S)}{8} = 1 \mu A$. Note that since the 8 levels to be resolved are in the Low Resistive State of the RRAM, the $G_{max}$ and $G_{min}$ are the minimum and maximum of the range in the LRS, which correspond to $40 \mu S$ and $120 \mu S$. 

In the \textbf{routing tile}, the read-out circuitry has to resolve between two levels which will either let the spike regenerate and thus propagate, or will get blocked. Therefore, the LSB of the $i_{in}$ current in the routing tile is $i_{in,{LSB}, R} = \frac{V_{ref} (G_{max}-G_{min})}{N}$. Based on the Fig.~\ref{fig:exp_neuron}d, this value for the neuron tile is $\frac{100 mV (40 \mu S - 10 \mu S)}{2} = 15 \mu A$. Note that since the 2 levels to be resolved are the LRS and HRS of the RRAM, the $G_{max}$ and $G_{min}$ correspond to $10 \mu S$ and $40 \mu S$. 

To be able to distinguish between any two levels in both cases, we will consider a maximum error of $\frac{i_{in,{LSB}}}{2}$. Therefore, the maximum tolerable error in the the neuron tile is $0.5 \mu A$ and in the routing tile is $7.5 \mu A$. 

This means that if the feedback loop does not close in $t_s$ of the pulse width, $V_x$ drop is a lot more tolerable in the routing tile than it is in the neuron tile. This suggests that the bandwidth requirements in the case of neuron tile is $7.5/0.5 = 15$ times more than that of the routing tile. The BW requirements directly translate to the biasing of the amplifier and thus its power consumption. Therefore, the static power consumption of the neuron tile is 15 times that of the routing tile.
The current requirements also translate to area, since larger currents require wider transistors. 

\section*{Appendix 3: Reconfigurable halide perovskite nanocrystal memristors for neuromorphic computing}

\paragraph{Supplementary Note 1} 

\begin{figure}[H]
  \centering
  \includegraphics[width=0.75\textwidth]{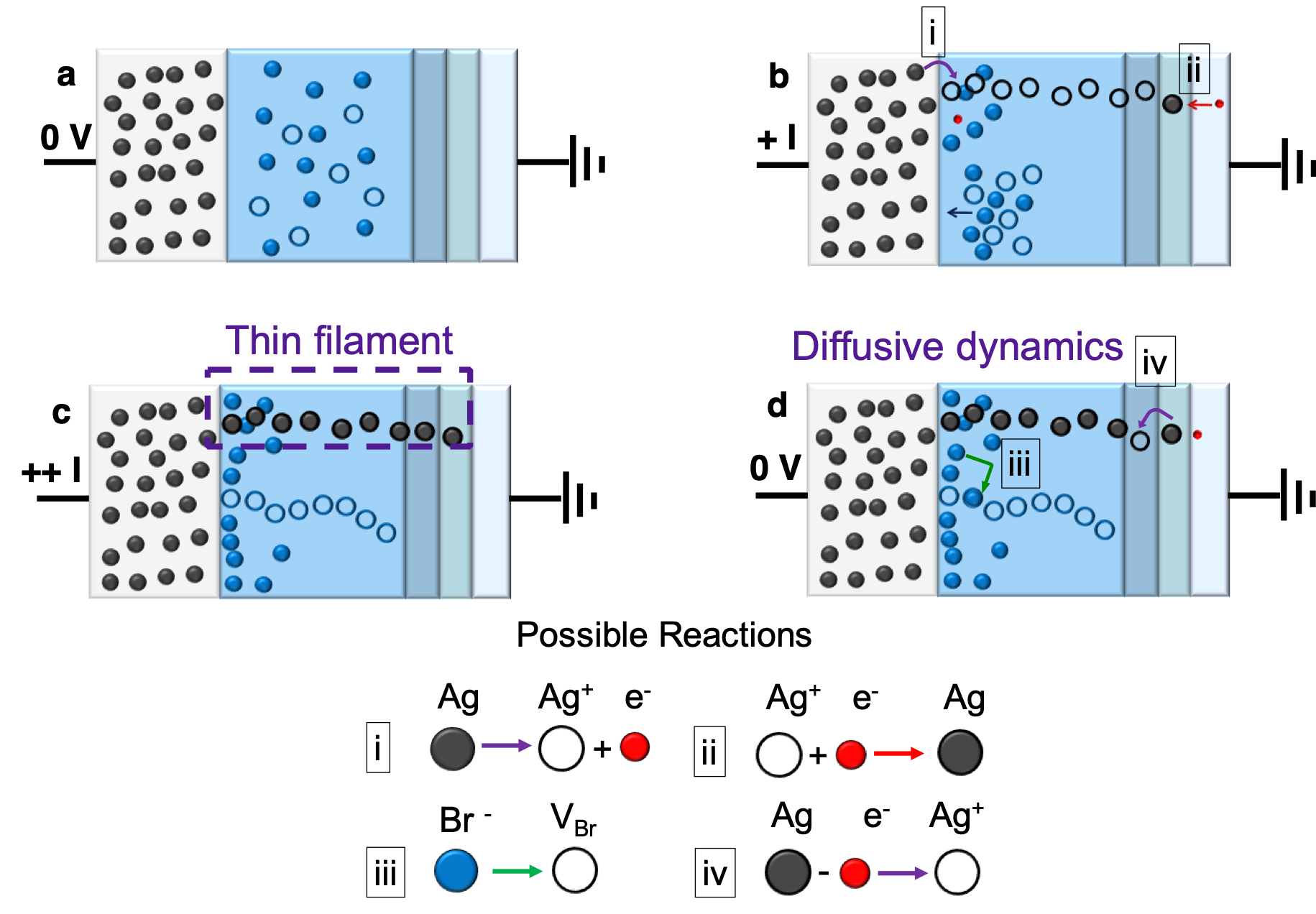}
  \caption{Proposed volatile diffusive switching mechanism. (a-d) illustrate the various stages of filament formation and rupture. i-iv indicate the possible reactions happening within the device.}
  \label{supfig:sup_fig4}
\end{figure}

Diffusive behaviour: The volatile threshold switching behaviour can be attributed to the redistribution of $\mathrm{Ag}^{+}$ and $\mathrm{Br}^{-}$ ions under an applied electric field, and their back-diffusion upon removing power. 
The soft lattice of the halide perovskite matrix has been observed to enable $\mathrm{Ag}^{+}$ migration with an activation energy $\sim 0.15 \mathrm{eV}^2$. 
Interestingly, migration of halide ions and their vacancies within the perovskite matrix also occur at similar energies $\sim 0.10-0.25 \mathrm{eV}$~\cite{John2018-kj,Zhu2017-pw, Nedelcu2015-gr}, making it difficult to pinpoint a singular operation mechanism. 
We hypothesize that during the SET process, $\mathrm{Ag}$ atoms are ionized to $\mathrm{Ag}^{+}$ and forms a percolation path through the device structure. 
Electrons from the grounded electrode oxidize $\mathrm{Ag}^{+}$to form weak $\mathrm{Ag}$ filaments. Parallelly, $\mathrm{Br}^{-}$ions get attracted towards the positively charged electrode and a weak filament composed of vacancies $\left(\mathrm{V}_{\mathrm{Br}}\right)$ is formed. 
Both these factors increase the device conductance from a high resistance state (HRS) to a temporary low resistance state (LRS). 
Upon removing the electric field, the low activation energy of the ions causes them to diffuse back spontaneously, breaking the percolation path and leading to volatile memory characteristics a.k.a. short-term plasticity. 
The low compliance current $\left(\mathrm{I}_{\mathrm{cc}}\right)$ of $1 \mu \mathrm{A}$ ensures that the electrochemical reactions are well regulated, and the percolation pathways formed are weak enough to allow these diffusive dynamics.

\paragraph{Supplementary Note 2} 

\begin{figure}[H]
  \centering
  \includegraphics[width=0.75\textwidth]{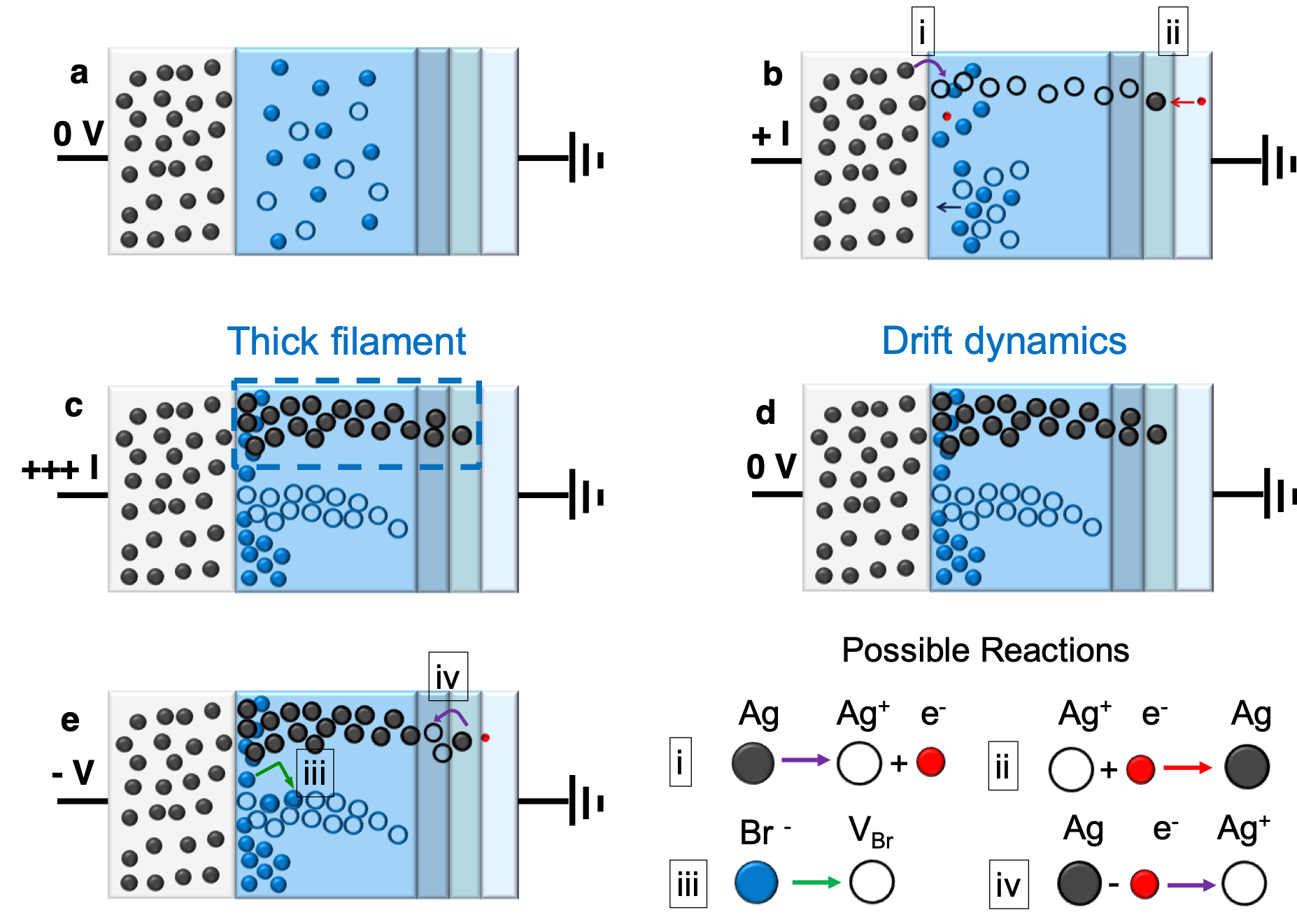}
  \caption{Proposed non-volatile drift switching mechanism. (a-e) illustrate the various stages of filament formation and rupture. i-iv indicate the possible reactions happening within the device.}
  \label{supfig:sup_fig6}
\end{figure}

\begin{figure}[H]
  \centering
  \includegraphics[width=0.35\textwidth]{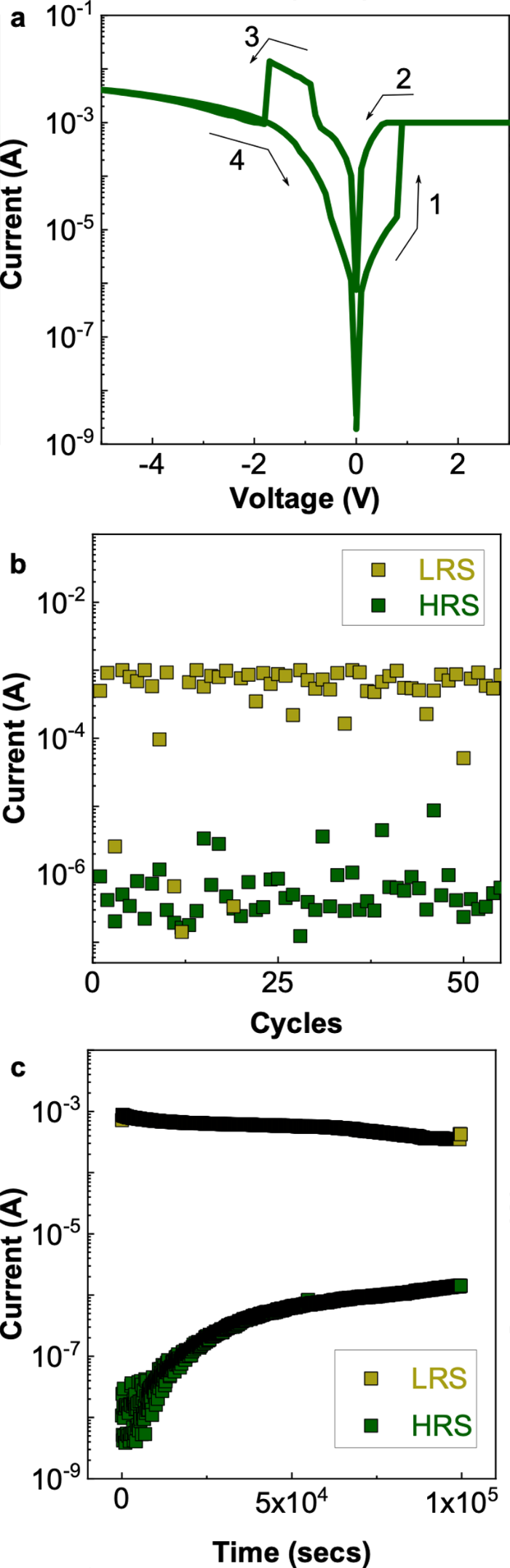}
  \caption{Non-volatile drift switching of DDAB-capped CsPbBr3 NC memristors. (a) Representative IV characteristics. (b) Endurance. (c) Retention.}
  \label{supfig:sup_fig7}
\end{figure}

\begin{figure}[H]
  \centering
  \includegraphics[width=0.5\textwidth]{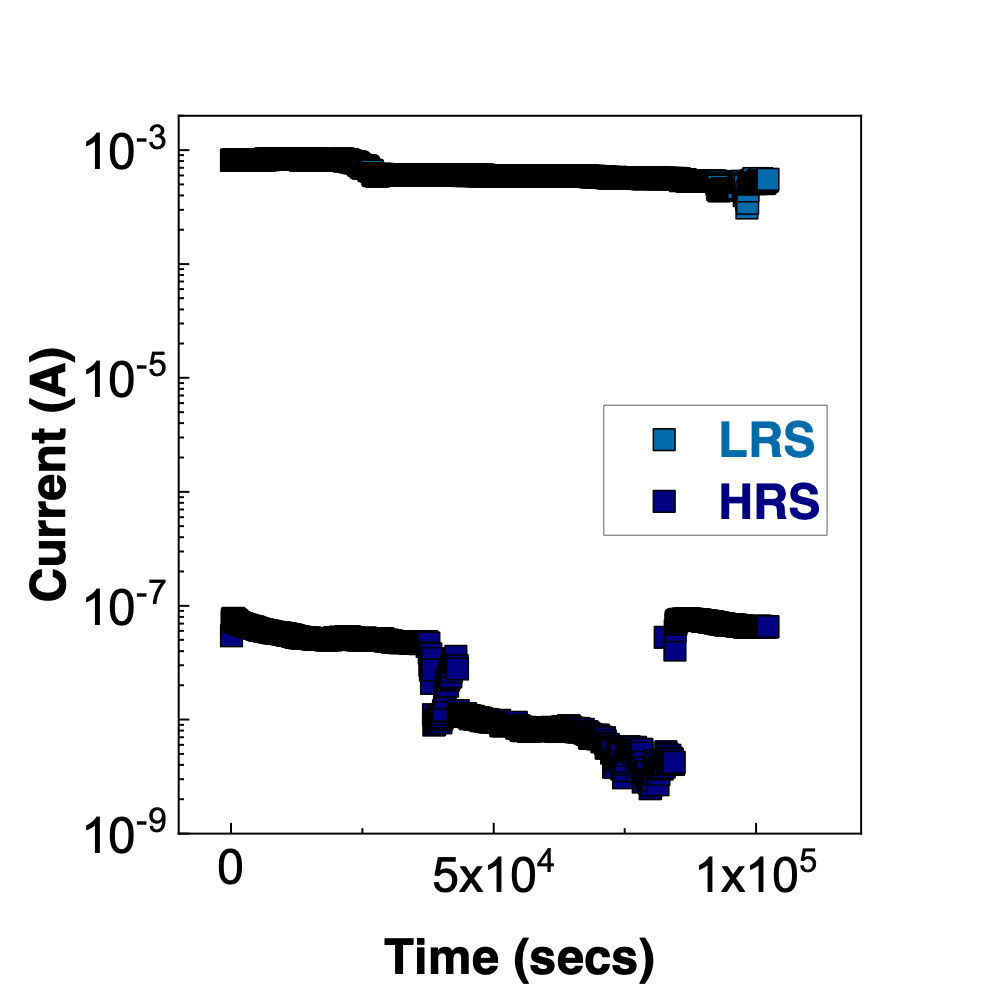}
  \caption{Non-volatile drift switching of OGB-capped CsPbBr3 NC memristors. Figure shows the retention performance.}
  \label{supfig:sup_fig8}
\end{figure}

Drift behaviour: Upon increasing the $\mathrm{I}_{\mathrm{cc}}$ to $1 \mathrm{~mA}$ (3 orders of magnitude higher than that used for volatile threshold switching), permanent and thicker conductive filamentary pathways are possibly formed within the device as illustrated in Fig. \ref*{supfig:sup_fig6}. 
This increases the device conductance from a high resistance state (HRS) to a permanent and much lower low resistance state (LRS). 
Electrochemical reactions are triggered to a higher extent and hence, the switching dynamics is now dominated by the drift kinetics of the mobile ion species $\mathrm{Ag}^{+}$ and $\mathrm{Br}^{-}$, rather than diffusion. 
Hence upon removing the electric field, the conductive filaments remain largely unaffected, and the devices retain their LRS and portray longterm plasticity. 
Application of voltage sweeps, or pulses of opposite polarity causes rupture of these filaments, and the devices are reset to their HRS. 
For DDAB-capped $\mathrm{CsPbBr}_3 \mathrm{NCs}$, the devices transition to a non-erasable non-volatile state within $\sim 50$ cycles, indicating formation of very thick filaments (Fig.~\ref{supfig:sup_fig7})
On the other hand, the OGB-capped $\mathrm{CsPbBr}_3$ NCs display a record-high nonvolatile endurance of 5655 cycles and retention of $10^5$ seconds (Fig.~\ref{supfig:sup_fig8}), pointing to better regulation of the filament formation and rupture kinetics.

\paragraph{Supplementary Note 3} 

\begin{figure}[H]
  \centering
  \includegraphics[width=0.35\textwidth]{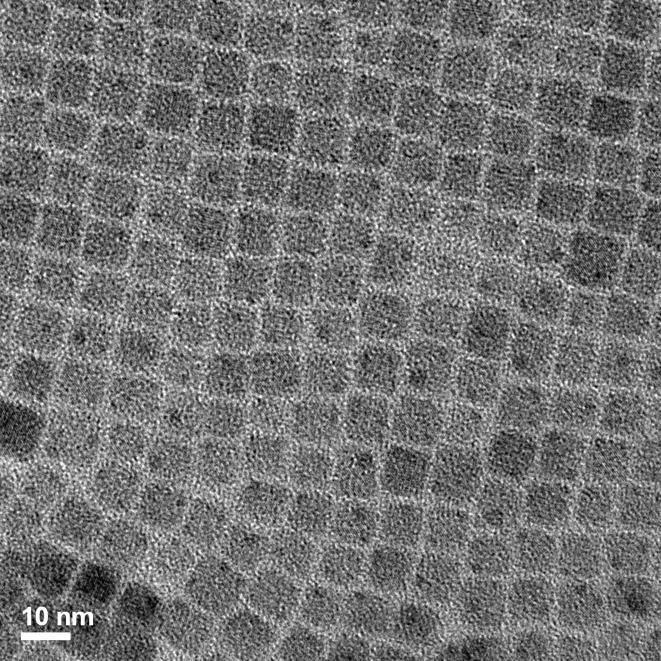}
  \caption{Transmission electron microscope (TEM) images of DDAB-capped CsPbBr3 NCs.}
  \label{supfig:sup_fig10}
\end{figure}

\paragraph{Supplementary Note 4} 

\begin{figure}[H]
  \centering
  \includegraphics[width=\textwidth]{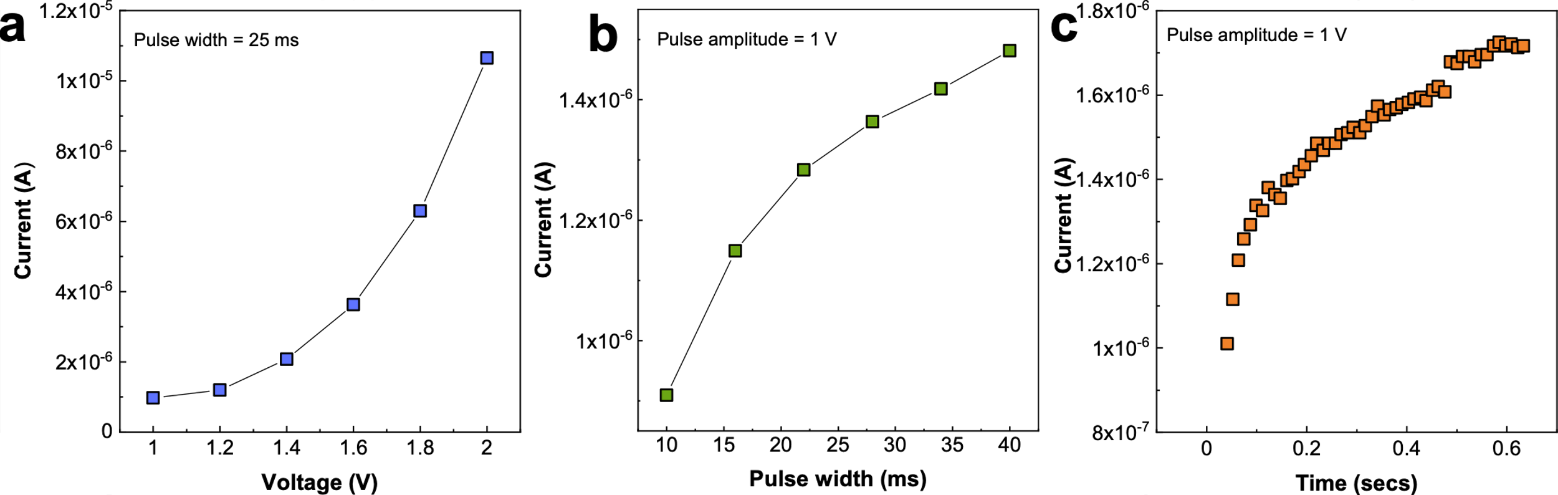}
  \caption{Non-linear variation of the device conductance as a function of the stimulation pulse (a) amplitude, (b) width and (c) number. For (a), the pulse width and number is kept constant at $25 \mathrm{~ms}$ and 1 respectively. For (b), the pulse amplitude and number is kept constant at $1 \mathrm{~V}$ and 1 respectively. For (c), the pulse amplitude and width is kept constant at $1 \mathrm{~V}$ and $25 \mathrm{~ms}$ respectively.}
  \label{supfig:sup_fig22}
\end{figure}

\begin{figure}[H]
  \centering
  \includegraphics[width=\textwidth]{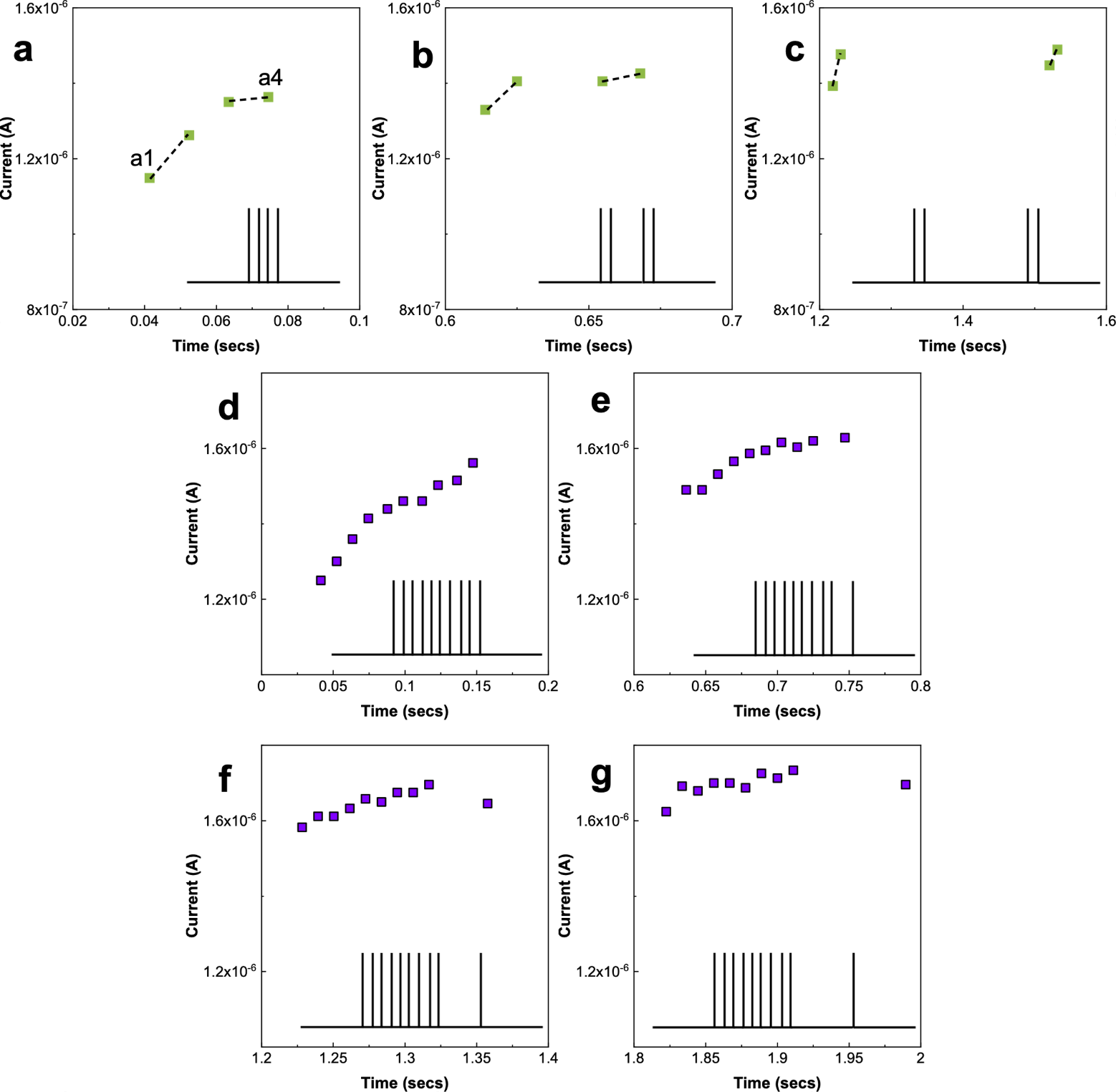}
  \caption{Echo state Properties. Variation in the device conductance of the volatile diffusive perovskite memristor as a function of the inter-group pulse interval. The interval between the two sequences increases from (a) $10 \mathrm{~ms}$, (b) $30 \mathrm{~ms}$ to (c) $300 \mathrm{~ms}$. (d-g) Current responses when subjected to 10 identical stimulation pulses $(1 \mathrm{~V}, 5 \mathrm{~ms})$ with different pulse interval conditions for the final pulse. The interval varies from (d) $10 \mathrm{~ms}$, (e) $23 \mathrm{~ms}$, (f) $41 \mathrm{~ms}$, to (g) $80 \mathrm{~ms}$.}
  \label{supfig:sup_fig23}
\end{figure}

The echo state property of a reservoir refers to the impact that previous inputs have on the current reservoir state, and how that influence fades out with time. 
To test this, four short pulses of $1 \mathrm{~V}, 5 \mathrm{~ms}$ are applied to the device in a paired-pulse format and the device states are recorded. 
A non-linear accumulative behaviour is observed as a function of the paired-pulse interval. 
In Fig~\ref{supfig:sup_fig23}a, a short paired-pulse interval of $10 \mathrm{~ms}$ results in an echo index (defined as $\left.\left(\frac{a 4}{a 1}\right) * 100\right)$ of $118 \%$. 
Longer intervals $(30 \mathrm{~ms}$ and $300 \mathrm{~ms}$ in Fig.~\ref{supfig:sup_fig23}b,c) result in smaller echo indices ( $107.5 \%$ and $107.2 \%$ respectively), reflective of the short-term memory in the perovskite memristors. 
To further test the echo state property, three pulse trains consisting of 10 identical stimulation pulses $(1 \mathrm{~V}, 5 \mathrm{~ms})$ are applied to the device and the device states are recorded. 
In all cases, a non-linear accumulative behaviour is observed. 
As shown in Fig.~\ref{supfig:sup_fig23}d-g, short intervals ( $\leq 23 \mathrm{~ms}$ ) for the last stimulation pulse result in further accumulation while long intervals result in depression of the device state. 
This indicates that the present device state remembers the input temporal features in the recent past but not the far past, allowing the diffusive perovskite memristors to act as efficient reservoir elements. 

\paragraph{Supplementary Note 5}

\begin{figure}[H]
  \centering
  \includegraphics[width=0.5\textwidth]{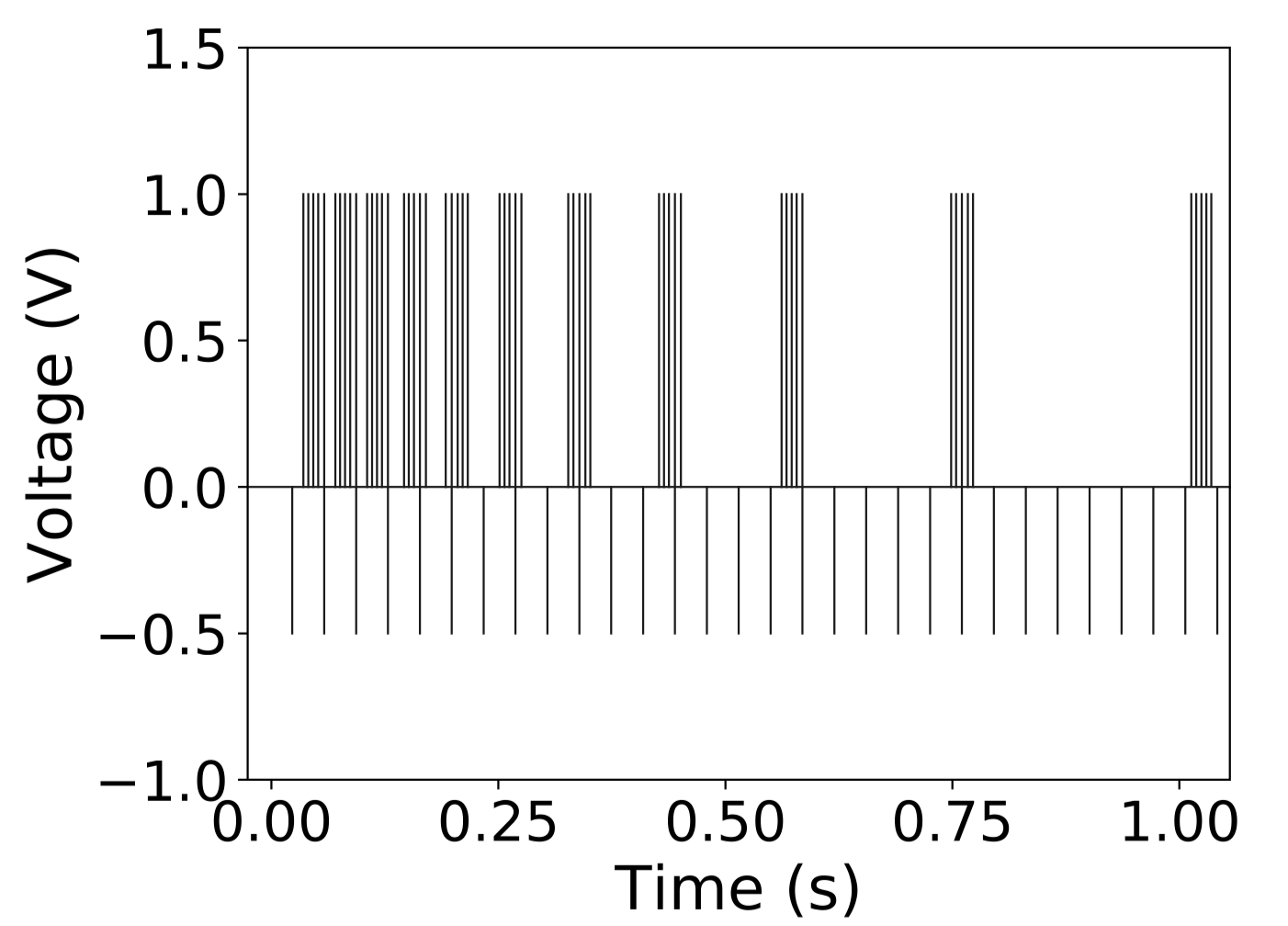}
  \caption{Input waveforms. A representative "Write" (amplitude $=1 \mathrm{~V}$, pulse width $=20$ $\mathrm{ms})$ and "Read" (amplitude $=-0.5 \mathrm{~V}$, pulse width $=5 \mathrm{~ms}$ ) spike train applied to the volatile perovskite memristors in the reservoir layer.}
  \label{supfig:sup_fig25}
\end{figure}

\begin{figure}[H]
  \centering
  \includegraphics[width=0.75\textwidth]{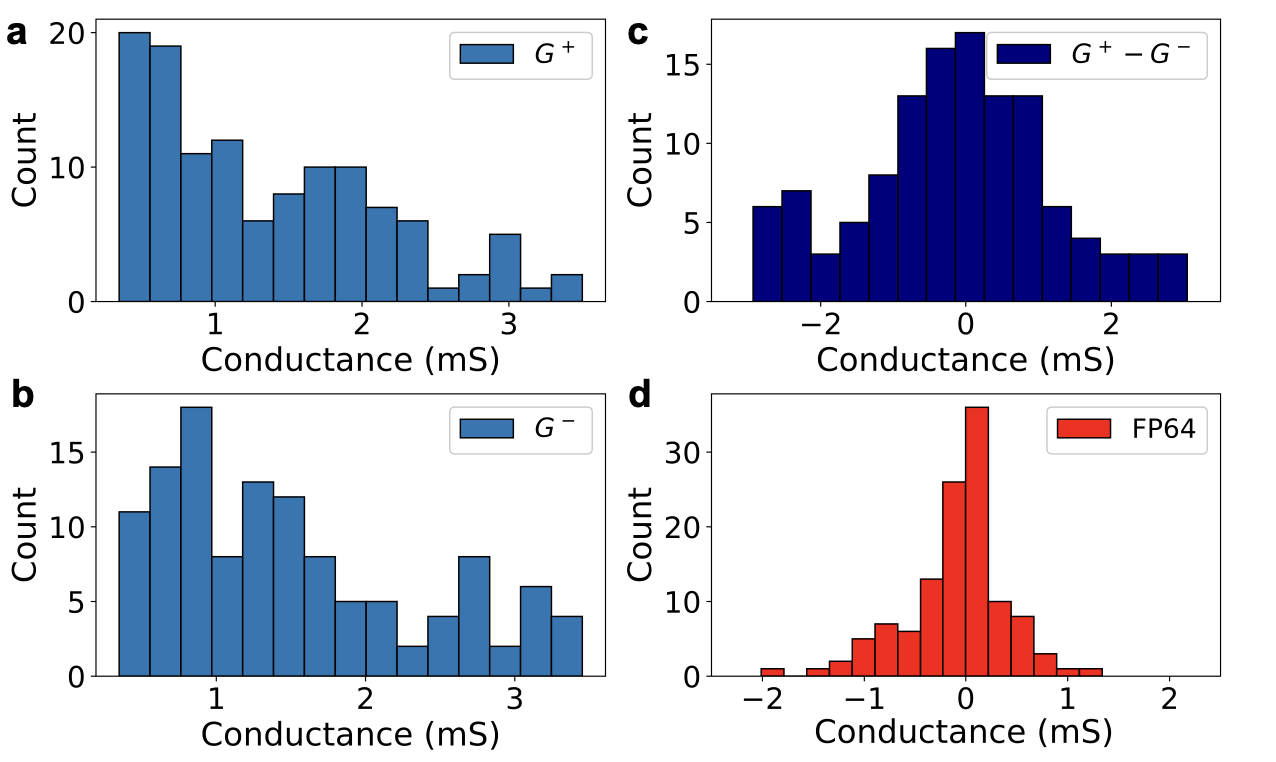}
  \caption{Weight distribution. The synaptic weight distribution after training. (a-b) Conductance distribution of both positive and negative differential perovskite memristors. (c) The effective (unscaled) conductance distribution. (d) Weight distribution of the readout layer with floating-point-double precision weights. The effective memristive weights and FP64 weights follow a similar distribution.}
  \label{supfig:sup_fig26}
\end{figure}

\begin{figure}[H]
  \centering
  \includegraphics[width=0.5\textwidth]{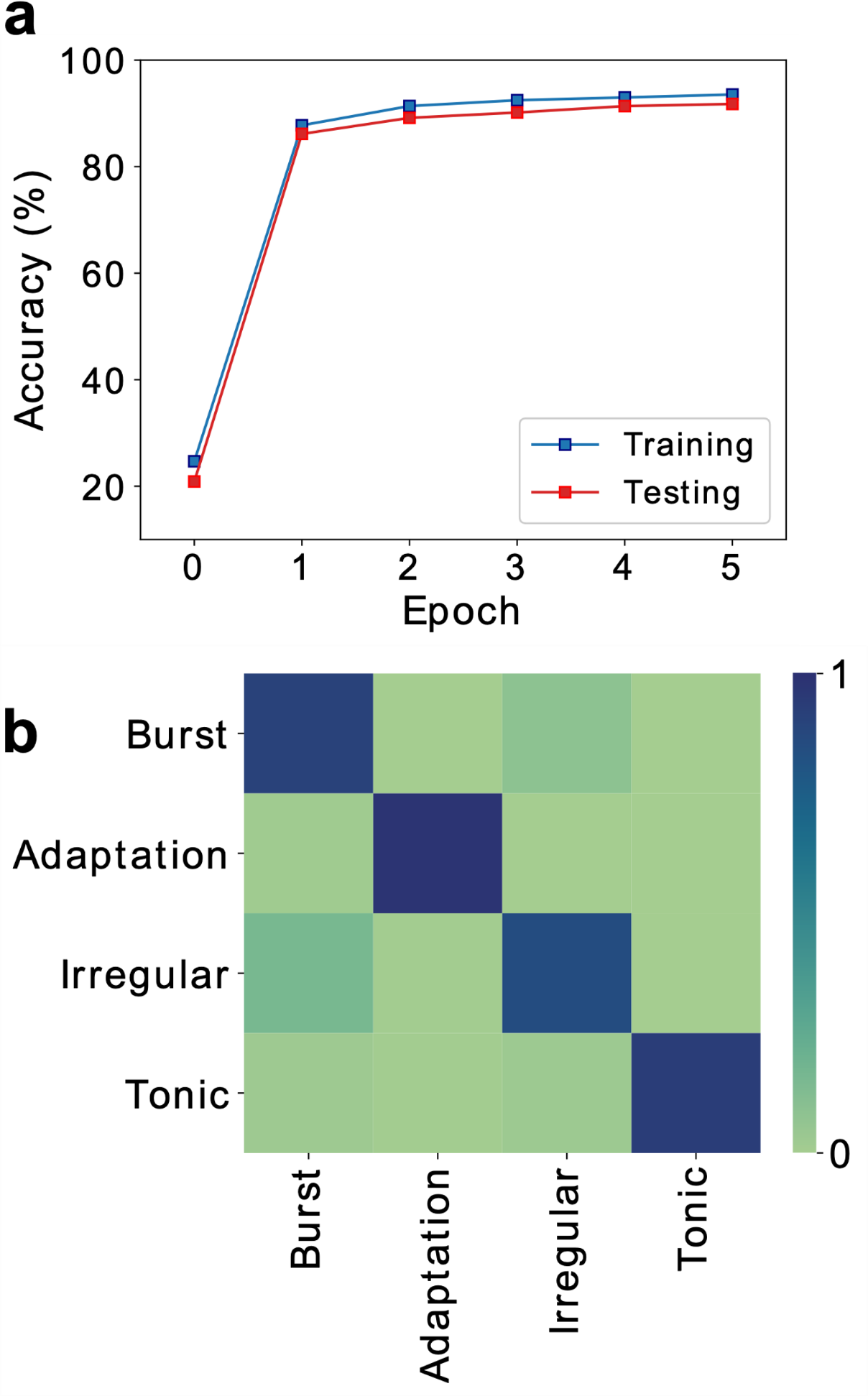}
  \caption{Training RC with FP64 readout weights using backpropagation. The training metrics of ANN is shown. (a) Training and testing accuracy over 5 epochs demonstrates that the network solves the classification task with high accuracy without overfitting. (b) Confidence matrix calculated at the end of training. The correct response probability is shown in the right color scale. It is evident that network performs slightly worse in discriminating irregular patterns.}
  \label{supfig:sup_fig27}
\end{figure}

\begin{table}[H]
  \centering
  \caption{Comparing training and test accuracies of both approaches. The neural spiking pattern classification performance table comparing the two approaches. The readout layer with drift-based halide-perovskite memristor weights trained with online $\left(\mathrm{I}_{\mathrm{cc}}\right)$ control achieves comparable result with FP64 weights trained with backpropagation.}
  \begin{tabular}{llcccccc}
  \toprule
   & & \multicolumn{6}{c}{Accuracy (\%)} \\
  \cmidrule(lr){3-8} 
   & & Epoch 0 & Epoch 1 & Epoch 2 & Epoch 3 & Epoch 4 & Epoch 5 \\
  \midrule
  \multirow{2}{*}{\begin{tabular}[c]{@{}l@{}}FP64 \\ with Backpropagation\end{tabular}} 
   & Training & 24.68 & 87.76 & 91.38 & 92.47 & 92.99 & 93.54 \\
   & Testing & 20.88 & 86.14 & 89.16 & 90.16 & 91.37 & 91.77 \\
  \midrule
  \multirow{2}{*}{\begin{tabular}[c]{@{}l@{}}Perovskite \\ with I\textsubscript{cc} Control\end{tabular}} 
   & Training & 10.32 & 69.11 & 89.12 & 83.09 & 85.59 & 86.75 \\
   & Testing & 14.46 & 73.29 & 86.14 & 80.92 & 84.94 & 85.14 \\
  \bottomrule
  \end{tabular}
  \label{table:sup_perovskite_table}
  \end{table}

  \begin{figure}[H]
    \centering
    \includegraphics[width=\textwidth]{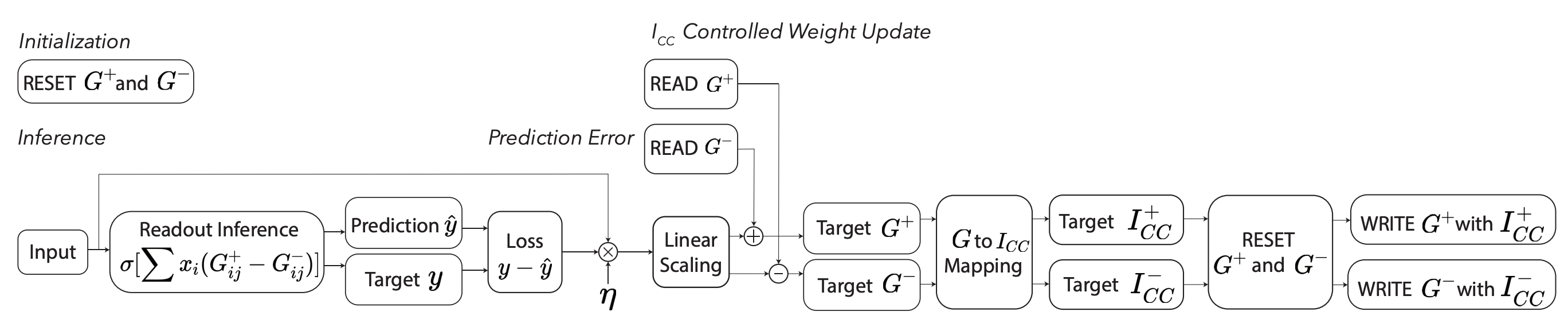}
    \caption{I $_{\mathbf{c c}}$ modulated training for drift-based perovskite configuration.}
    \label{supfig:sup_fig28}
  \end{figure}

  During the inference procedure, reservoir output vector of length 30 is fed into the readout layer. Memristors in the readout layer are placed in the differential architecture, in which the difference of conductance values of two differential memristors $\left(G^{+}\right.$and $\left.G^{-}\right)$determines the effective synaptic strength. 
  Scaled with $\beta=1 /\left(G_{\max }-G_{\min }\right)$, where $G_{\max }=0.35 \mathrm{mS}$ and $G_{\min }=0.1 \mathrm{mS}$, the weight matrix $(30 \times 4)$ is calculated as $W=\beta\left[G^{+}-G^{-}\right]$. And the network prediction is determined by choosing output neuron index with the maximum activation level.
  
  For the training procedure, the network loss is calculated as the difference between output layer prediction and the one-hot encoded target vector indicating one of the four firing patterns. 
  At this point, one can calculate targeted weights using the backpropagation algorithm. However, to support fully online-learning, we tested $\mathrm{I}_{\mathrm{cc}}$ controlled weight update scheme where following stages in the pipeline can be easily implemented with the mixed-signal circuits in an event-driven manner. The $\mathrm{I}_{\mathrm{cc}}$ controlled weight update is implemented as follows. 
  First, the required weight change is calculated with $W_{\text {target }}=\eta x_i \delta_j$, where $\eta$ is suitably low learning rate, $x_i$ is the reservoir layer output and $\delta_j$ is the calculated error. In order to calculate the target conductance values for both positive and negative memristors, we first linearly scale the weight change to conductance change (by multiplying with $1 / \beta$ ). 
  Secondly, we read both the positive and negative conductance values. 
  By using a push-pull mechanism, we calculate the target conductance values. 
  The push-pull mechanism ensures a higher dynamic range in the differential configuration. 
  Third, the target conductances are linearly mapped to the target $\mathrm{I}_{\mathrm{cc}}$ values $\left(\mathrm{I}_{\mathrm{cc}, \text { target }}=\left(G_{\text {target }}+1.249 \times 10^{-5}\right) /\right.$ 3.338) for positive and negative memristors. 
  The weights are updated with the application of RESET and SET pulses with the targeted $\mathrm{I}_{\mathrm{cc}}$. 
  Using the linear relation between $\mathrm{I}_{\mathrm{cc}} \rightarrow G$ control, we calculated mean $\left(\mu_G=3.338 \mathrm{I}_{\mathrm{cc}}-1.294 \times 10^{-5}\right)$ and standard deviation $\sigma_G=7.040 \mathrm{I}_{\mathrm{cc}}+3.0585$ and sample from the corresponding Normal distribution.)

  \paragraph{Supplementary Note 6}
  \begin{figure}[H]
    \centering
    \includegraphics[width=\textwidth]{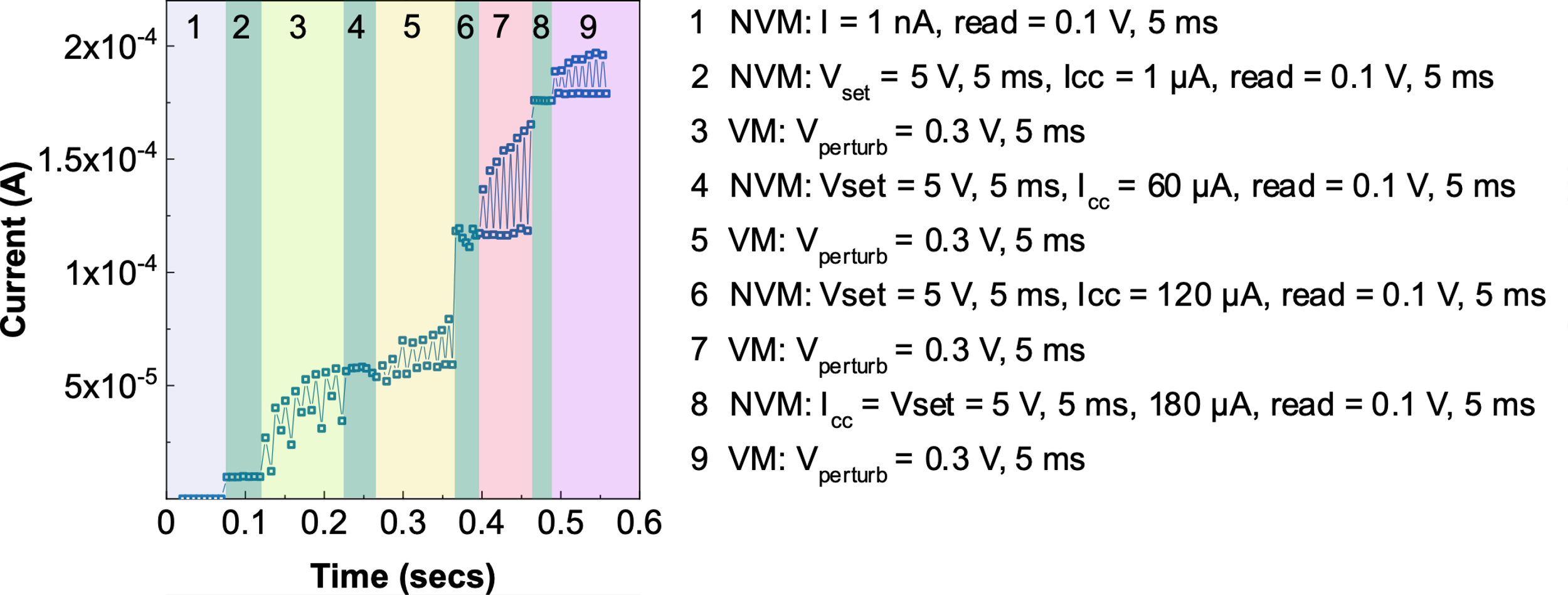}
    \caption{“Reconfigurability” on-the-fly of OGB-capped CsPbBr 3NC memristors. The device is switched between its non-volatile and volatile mode on demand.}
    \label{supfig:sup_fig29}
  \end{figure}

  To demonstrate "reconfigurability" on-the-fly, our devices are switched between volatile and non-volatile modes with precise compliance current ( $\mathrm{I}_{\mathrm{cc}}$ ) control and selection of activation voltages. Fig.~\ref{supfig:sup_fig29} shows that our devices can act as a volatile memory even after setting it to multiple non-volatile states. This proves true "reconfigurability" of our devices, hitherto undemonstrated. Such behaviour is an example of the neuromorphic implementation of synapses in Spiking Neural Networks (SNNs) that demand both volatile and non-volatile switching properties, simultaneously.

\cleardoublepage%
\def\dir{chapters/1_precision}

\cleardoublepage%
\def\dir{chapters/2_onlinelearning}

\cleardoublepage%
\def\dir{chapters/3_perovskite}

\cleardoublepage%
\def\dir{chapters/4_mosaic}

\cleardoublepage%
\def\dir{chapters/5_conclusions}

\appendix
\cleardoublepage%
\def\dir{chapters/6_appendix}

\cleardoublepage%********************************************************************
% Bibliography
%*******************************************************
% work-around to have small caps also here in the headline
\manualmark%
\markboth{\spacedlowsmallcaps{\bibname}}{\spacedlowsmallcaps{\bibname}} % work-around to have small caps also
%\phantomsection 
\refstepcounter{dummy}
\addtocontents{toc}{\protect\vspace{\beforebibskip}} % to have the bib a bit from the rest in the toc
\addcontentsline{toc}{chapter}{\tocEntry{\bibname}}
\label{app:bibliography}
{%
  \emergencystretch=1em%
  \printbibliography%
}

\bookmarksetup{startatroot}
%\pagenumbering{gobble}
%\cleardoublepage\include{frontbackmatter/cv}
\cleardoublepage\manualmark%
%\markboth{\spacedlowsmallcaps{\bibname}}{\spacedlowsmallcaps{\bibname}} % work-around to have small caps also
%\phantomsection 
\refstepcounter{dummy}
%\addtocontents{toc}{\protect\vspace{\beforebibskip}} % to have the bib a bit from the rest in the toc
\addcontentsline{toc}{chapter}{\tocEntry{Contributions}}
\label{app:contributions}

This thesis consists of six selected publications, conducted in collaboration with electrical engineers, computer scientists, material designers and neuroscientists.
In this section, I roughly discuss my personal contributions to each project.\\

\textbf{Analog weight updates with compliance current modulation of binary ReRAMs for on-chip learning} (Chapter~\ref{ch:precision})
\begin{itemize}[noitemsep, topsep=5pt]
\item collaborated with material scientists for data collection and modeling (e.g., Fig. 1)
\item coded training and evaluation of \ac{RRAM}-based neural network simulations (e.g., Alg. 1)
\item assisted circuit design with simulation findings (e.g., Fig. 2)
\item contributed to majority of the manuscript including figures
\end{itemize}

\vspace{0.4cm}
\textbf{Online training of spiking recurrent neural networks with Phase-Change Memory synapses} (Chapter~\ref{ch:onlinelearning})
\begin{itemize}[noitemsep, topsep=0pt]
\item performed literature review to identify the problem
\item coded e-prop, \ac{PCM}-based analog simulation framework, and neural network training
\item conducted all data analysis and visualization
\item contributed to majority of the manuscript including figures
\end{itemize}

\vspace{0.4cm}
\textbf{Biologically-inspired training of spiking recurrent neural networks with neuromorphic hardware} (Chapter~\ref{ch:onlinelearning})
\begin{itemize}[noitemsep, topsep=5pt]
\item collaboratively planned the project with IBM researchers and INI (e.g., work assignments, experiments, deadlines)
\item assisted with experiment datasets, architecture design following prior work~\cite{Demirag2021-xc}
\item weekly supervision of \text{Anja \v{S}urina} (e.g., debugging, hyperparameter opt...)
\item assisted paper writing and designed several figures
\end{itemize}

\vspace{0.4cm}
\textbf{PCM-trace: scalable synaptic eligibility traces with resistivity drift of Phase-Change Materials} (Chapter~\ref{ch:onlinelearning})
\begin{itemize}[noitemsep, topsep=5pt]
\item performed literature review to identify the problem
\item collaborated with material scientists for data collection and modeling
\item coded pcm-trace and multi-pcm trace experiments and analysis
\item assisted circuit design with simulation findings
\item contributed to majority of the manuscript 
\end{itemize}

\vspace{0.4cm}
\textbf{Reconfigurable halide perovskite nanocrystal memristors for neuromorphic computing} (Chapter~\ref{ch:perovskite})
\begin{itemize}[noitemsep, topsep=5pt]
\item performed literature review to identify the problem
\item collaborated with material scientists for data collection, required device specifications and modeling (e.g. non-volatility time constant, Fig.4b-c)
\item coded training and evaluation of simulations with volatile and non-volatile memristor models (e.g., RC framework)
\item designed $I_{CC}$ modulated training following prior work~\cite{Payvand2020-fu} (e.g., Supplementary Fig. 28)
\item contributed to majority of the manuscript
\end{itemize}

\vspace{0.4cm}
\textbf{Mosaic: in-memory computing and routing for small-world spike-based neuromorphic systems} (Chapter~\ref{ch:Mosaic})
\begin{itemize}[noitemsep, topsep=5pt]
\item performed extensive literature review to identify the strengths the idea
\item collaborated with material scientists for data collection and modeling (e.g., Fig. 2d)
\item coded layout-aware training and evaluation of \ac{SHD} with backprop and RL with \ac{ES} benchmarks (e.g., Fig. 4)
\item contributed to majority of the manuscript
\end{itemize}
\cleardoublepage%*******************************************************
% Publications
%*******************************************************
\manualmark%
%\markboth{\spacedlowsmallcaps{\bibname}}{\spacedlowsmallcaps{\bibname}} % work-around to have small caps also
%\phantomsection 
\refstepcounter{dummy}
%\addtocontents{toc}{\protect\vspace{\beforebibskip}} % to have the bib a bit from the rest in the toc
\addcontentsline{toc}{chapter}{\tocEntry{Publications}}
\label{app:publications}

%\pdfbookmark[1]{Publications}{publications}
%\chapter{Publications}

\noindent
Articles in peer-reviewed journals:
\begin{refsection}[ownpubs]
  \small%
  \nocite{*}
  \printbibliography[heading=none,type=article]
\end{refsection}

\noindent
Preprints:
\begin{refsection}[ownpubs]
  \small%
  \nocite{*}
  \printbibliography[heading=none,keyword=arxiv]
\end{refsection}

\noindent
Conference contributions:
\begin{refsection}[ownpubs]
  \small%
  \nocite{*}
  \printbibliography[heading=none,type=inproceedings]

@article{John2022-ph,
  title        = {Reconfigurable halide perovskite nanocrystal memristors for
                  neuromorphic computing},
  author       = {John, Rohit Abraham and Demirag, Yigit and Shynkarenko, Yevhen
                  and Berezovska, Yuliia and Ohannessian, Natacha and Payvand,
                  Melika and Zeng, Peng and Bodnarchuk, Maryna I and Krumeich,
                  Frank and Kara, Gökhan and Shorubalko, Ivan and Nair, Manu V
                  and Cooke, Graham A and Lippert, Thomas and Indiveri, Giacomo
                  and Kovalenko, Maksym V},
  journaltitle = {Nat. Commun.},
  publisher    = {Springer Science and Business Media LLC},
  volume       = {13},
  issue        = {1},
  pages        = {2074},
  date         = {2022-04-19},
  urldate      = {2023-08-12},
  language     = {en}
}

@article{Dalgaty2024-or,
  title        = {Mosaic: in-memory computing and routing for small-world
                  spike-based neuromorphic systems},
  author       = {Dalgaty, Thomas and Moro, Filippo and Demirag, Yigit and De
                  Pra, Alessio and Indiveri, Giacomo and Vianello, Elisa and
                  Payvand, Melika},
  journaltitle = {Nat. Commun.},
  publisher    = {Springer Science and Business Media LLC},
  volume       = {15},
  issue        = {1},
  pages        = {1--12},
  date         = {2024-01-02},
  urldate      = {2024-01-05},
  language     = {en}
}

@article{D-Agostino2024-bh,
  title        = {{DenRAM}: neuromorphic dendritic architecture with {RRAM} for
                  efficient temporal processing with delays},
  author       = {D'Agostino, Simone and Moro, Filippo and Torchet, Tristan and
                  Demirag, Yigit and Grenouillet, Laurent and Castellani,
                  Niccolò and Indiveri, Giacomo and Vianello, Elisa and Payvand,
                  Melika},
  journaltitle = {Nat. Commun.},
  publisher    = {Springer Science and Business Media LLC},
  volume       = {15},
  issue        = {1},
  pages        = {1--12},
  date         = {2024-04-24},
  urldate      = {2024-05-12},
  language     = {en}
}

@preprint{Demirag2021-xc,
  title        = {Online training of spiking recurrent neural networks with
                  Phase-Change Memory synapses},
  author       = {Demirag, Yigit and Frenkel, Charlotte and Payvand, Melika and
                  Indiveri, Giacomo},
  journaltitle = {arXiv [cs.NE]},
  date         = {2021-08-03},
  eprintclass  = {cs.NE}
}

@inproceedings{Demirag2021-kd,
  title      = {{PCM}-trace: Scalable synaptic eligibility traces with
                resistivity drift of phase-change materials},
  author     = {Demirag, Yigit and Moro, Filippo and Dalgaty, Thomas and
                Navarro, Gabriele and Frenkel, Charlotte and Indiveri, Giacomo
                and Vianello, Elisa and Payvand, Melika},
  booktitle  = {2021 IEEE International Symposium on Circuits and Systems
                (ISCAS)},
  publisher  = {IEEE},
  eventtitle = {2021 IEEE International Symposium on Circuits and Systems
                (ISCAS)},
  venue      = {Daegu, Korea},
  volume     = {00},
  pages      = {1--5},
  date       = {2021-05}
}

@inproceedings{Demirag2024-og,
  title     = {Network of biologically plausible neuron models can solve motor
               tasks through heterogeneity},
  author    = {Demirag, Yigit and Indiveri, Giacomo},
  booktitle = {Computational and Systems Neuroscience (COSYNE)},
  location  = {Lisbon, Portugal},
  date      = {2024-03},
}

@inproceedings{Demirag2023-ln,
  title     = {Overcoming phase-change material non-idealities by meta-learning
               for adaptation on the edge},
  author    = {Demirag, Yigit and Dittmann, Regina and Indiveri, Giacomo and
               Neftci, Emre},
  booktitle = {Proceedings of Neuromorphic Materials, Devices, Circuits and
               Systems (NeuMatDeCaS)},
  date      = {2023-01-09},
  urldate   = {2023-09-09}
}

@inproceedings{Payvand2020-fu,
  title      = {Analog weight updates with compliance current modulation of
                binary {ReRAMs} for on-chip learning},
  author     = {Payvand, Melika and Demirag, Yigit and Dalgaty, Thomas and
                Vianello, Elisa and Indiveri, Giacomo},
  booktitle  = {2020 IEEE International Symposium on Circuits and Systems
                (ISCAS)},
  publisher  = {IEEE},
  eventtitle = {2020 IEEE International Symposium on Circuits and Systems
                (ISCAS)},
  venue      = {Seville, Spain},
  volume     = {00},
  pages      = {1--5},
  date       = {2020-10}
}

@inproceedings{Bohnstingl2022-il,
  title      = {Biologically-inspired training of spiking recurrent neural
                networks with neuromorphic hardware},
  author     = {Bohnstingl, Thomas and Surina, Anja and Fabre, Maxime and
                Demirag, Yigit and Frenkel, Charlotte and Payvand, Melika and
                Indiveri, Giacomo and Pantazi, Angeliki},
  booktitle  = {2022 IEEE 4th International Conference on Artificial
                Intelligence Circuits and Systems (AICAS)},
  publisher  = {IEEE},
  eventtitle = {2022 IEEE 4th International Conference on Artificial
                Intelligence Circuits and Systems (AICAS)},
  venue      = {Incheon, Korea, Republic of},
  volume     = {00},
  pages      = {218--221},
  date       = {2022-06-13}
}

@inproceedings{Payvand2023-sl,
  title      = {Dendritic computation through exploiting resistive memory as
                both delays and weights},
  author     = {Payvand, Melika and D'Agostino, Simone and Moro, Filippo and
                Demirag, Yigit and Indiveri, Giacomo and Vianello, Elisa},
  booktitle  = {Proceedings of the 2023 International Conference on Neuromorphic
                Systems},
  location   = {New York, USA},
  eventtitle = {International Conference on Neuromorphic Systems (ICONS)},
  pages      = {1--4},
  date       = {2023-08}
}

@inproceedings{Raghunathan2024-gl,
  title     = {Hardware-aware Few-shot Learning on a Memristor-based Small-world
               Architecture},
  author    = {Raghunathan, Karthik Charan and Demirag, Yigit and Neftci, Emre
               and Payvand, Melika},
  booktitle = {Neuro Inspired Computational Elements Conference (NICE)},
  publisher = {IEEE},
  date      = {2024-06}
}

@inproceedings{Raghunathan2023Poster,
  author = {Raghunathan, Karthik Charan and Demirag, Yigit and Moro, Filippo and Neftci, Emre and Payvand, Melika},
  title = {Few-shot learning on brain-inspired small-world graphical hardware},
  booktitle = {International Conference on Neuromorphic, Natural and Physical Computing (NNPC 2023)},
  year = {2023},
  month = {October},
  address = {Hannover, Germany},
}

@inproceedings{Yu2021-ii,
  title      = {{BEOL} compatible cross-bar array of ferroelectric synapses},
  author     = {Yu, Zhenming and Bégon-Lours, Laura and Demirag, Yigit and Offrein,
                Bert Jan},
  booktitle  = {Proceedings of the 2021 International Conference on Neuromorphic
                Systems},
  eventtitle = {International Conference on Neuromorphic Systems (ICONS)},
  date       = {2021-06}
}

@inproceedings{Yik2023-hc,
  title        = {{NeuroBench}: Advancing neuromorphic computing through
                  collaborative, fair and representative benchmarking},
  author       = {Yik, Jason and Ahmed, Soikat Hasan and Ahmed, Zergham and
                  Anderson, Brian and Andreou, Andreas G and Bartolozzi, Chiara
                  and Basu, Arindam and Blanken, Douwe den and Bogdan, Petrut
                  and Bohte, Sander and Bouhadjar, Younes and Buckley, Sonia and
                  Cauwenberghs, Gert and Corradi, Federico and de Croon, Guido
                  and Danielescu, Andreea and Daram, Anurag and Davies, Mike and
                  Demirag, Yigit and Eshraghian, Jason and Forest, Jeremy and
                  Furber, Steve and Furlong, Michael and Gilra, Aditya and
                  Indiveri, Giacomo and Joshi, Siddharth and Karia, Vedant and
                  Khacef, Lyes and Knight, James C and Kriener, Laura and
                  Kubendran, Rajkumar and Kudithipudi, Dhireesha and Lenz,
                  Gregor and Manohar, Rajit and Mayr, Christian and Michmizos,
                  Konstantinos and Muir, Dylan and Neftci, Emre and Nowotny,
                  Thomas and Ottati, Fabrizio and Ozcelikkale, Ayca and
                  Pacik-Nelson, Noah and Panda, Priyadarshini and Pao-Sheng, Sun
                  and Payvand, Melika and Pehle, Christian and Petrovici, Mihai
                  A and Posch, Christoph and Renner, Alpha and Sandamirskaya,
                  Yulia and Schaefer, Clemens J S and van Schaik, André and
                  Schemmel, Johannes and Schuman, Catherine and Seo, Jae-Sun and
                  Sheik, Sadique and Shrestha, Sumit Bam and Sifalakis, Manolis
                  and Sironi, Amos and Stewart, Kenneth and Stewart, Terrence C
                  and Stratmann, Philipp and Tang, Guangzhi and Timcheck,
                  Jonathan and Verhelst, Marian and Vineyard, Craig M and
                  Vogginger, Bernhard and Yousefzadeh, Amirreza and Zhou, Biyan
                  and Zohora, Fatima Tuz and Frenkel, Charlotte and Reddi, Vijay
                  Janapa},
  journaltitle = {arXiv [cs.AI]},
  date         = {2023-04-10},
  eprintclass  = {cs.AI}
}
\end{refsection}

\end{document}